\documentclass{article}

\usepackage[preprint, nonatbib]{neurips_2023}

\usepackage[utf8]{inputenc} 
\usepackage[T1]{fontenc}    
\usepackage[hidelinks]{hyperref}       
\usepackage{url}            
\usepackage{booktabs}       
\usepackage{amsfonts}       
\usepackage{nicefrac}       
\usepackage{microtype}      
\usepackage{xcolor}         

\usepackage{quaternionCommands}
\usepackage{colors}
\usepackage{amsmath}
\usepackage{float}
\usepackage{soul}
\usepackage{caption}
\usepackage{subcaption}
\usepackage[printonlyused, nolist]{acronym}
\usepackage{multicol}
\usepackage{multirow}
\usepackage{makecell}
\usepackage[final]{changes}  %
\usepackage{mathtools}
\usepackage{amsthm}
\usepackage{enumitem} 
\usepackage[sorting=none]{biblatex}
\DeclareSourcemap{
  \maps[datatype=bibtex]{
    \map{
      \step[fieldsource=eprinttype, final]
      \step[fieldset=urldate, null]
      \step[fieldset=url, null]
    }
    \map{
      \pernottype{online}
      \pernottype{techreport}
      \step[fieldset=urldate, null]
      \step[fieldset=url, null]
    }
  }
}

\usepackage{tikz}
\usepackage{pgfplots}
\pgfplotsset{compat=newest}
\usepackage{pgfplotstable}
\usepgfplotslibrary{fillbetween}
\usetikzlibrary{backgrounds,calc,shadings,shapes.arrows,arrows,shapes.symbols,shadows,positioning,decorations.markings,backgrounds,arrows.meta}

\usepackage{pifont}
\newcommand{\cmark}{\ding{51}}%
\newcommand{\xmark}{\ding{55}}%

\DeclareMathOperator{\tanhshrink}{tanhshrink}
\DeclareMathOperator{\atan2}{atan2}

\newtheorem*{remark}{Remark}

\title{Improving Quaternion Neural Networks with Quaternionic Activation Functions}

\author{%
  Johannes Pöppelbaum\\
  Department of Automation Technology and Learning Systems\\
  South Westphalia University of Applied Sciences\\
  Soest, Germany \\
  \texttt{poeppelbaum.johannes@fh-swf.de} \\
  \And
  Andreas Schwung\\
  Department of Automation Technology and Learning Systems\\
  South Westphalia University of Applied Sciences\\
  Soest, Germany \\
  \texttt{schwung.andreas@fh-swf.de} \\
}

\begin{document}


\maketitle

\begin{abstract}
    In this paper, we propose novel quaternion activation functions where we modify either the quaternion magnitude or the phase, as an alternative to the commonly used split activation functions. We define criteria that are relevant for quaternion activation functions, and subsequently we propose our novel activation functions based on this analysis.
Instead of applying a known activation function like the ReLU or Tanh on the quaternion elements separately, these activation functions consider the quaternion properties and respect the quaternion space $\mathbb{H}$.
In particular, all quaternion components are utilized to calculate all output components, carrying out the benefit of the Hamilton product in e.g. the quaternion convolution to the activation functions.
The proposed activation functions can be incorporated in arbitrary quaternion valued neural networks trained with gradient descent techniques.
We further discuss the derivatives of the proposed activation functions where we observe beneficial properties for the activation functions affecting the phase.
Specifically, they prove to be sensitive on basically the whole input range, thus improved gradient flow can be expected.
We provide an elaborate experimental evaluation of our proposed quaternion activation functions including comparison with the split ReLU and split Tanh on two image classification tasks using the CIFAR-10 and SVHN dataset. There, especially the quaternion activation functions affecting the phase consistently prove to provide better performance.
\end{abstract}

\begin{acronym}[DQLL]
    \acro{NN}{neural network}
    \acro{QNN}{quaternion neural network}
\end{acronym}
\newcommand{\magPlot}[1]{%
\begin{tikzpicture}[tight background]%
\begin{axis}[%
        colormap/viridis,
        xlabel=Re($\quaternion{z}$),
        ylabel=$\lVert\text{Im}(\quaternion{z})\rVert$,
        zlabel=$\lVert \quaternion{f}(\quaternion{z}) \rVert$,
        xlabel style = {sloped, anchor=center},
        ylabel style = {sloped, anchor=center},
        ylabel shift = 3pt,
        zlabel shift = -3pt,
        grid=major,
        axis lines*=left,
        view={-45}{10},
        tick style={draw=none},
        ticklabel style = {font=\scriptsize},
        xtick={-1, -0.5, 0, 0.5, 1},
        ytick={0, 0.25, 0.5, 0.75, 1},
        ztick={0, 0.5, 1, 1.5},
        width=\linewidth,
        height=0.74\linewidth,
    ]%
    \addplot3 [surf, ultra thin, line width=0.1pt] table[x=re, y=im, z=norm, col sep=comma] {#1};%
\end{axis}%
\end{tikzpicture}%
}%
\newcommand{\magPlotTwo}[1]{%
\begin{tikzpicture}[tight background]%
\begin{axis}[%
        colormap/viridis,
        xlabel=Re($\quaternion{z}$),
        ylabel=$\lVert\text{Im}(\quaternion{z})\rVert$,
        zlabel=$\lVert \quaternion{f}(\quaternion{z}) \rVert$,
        xlabel style = {sloped, anchor=center},
        ylabel style = {sloped, anchor=center},
        ylabel shift = 3pt,
        zlabel shift = -3pt,
        grid=major,
        axis lines*=left,
        view={-45}{10},
        tick style={draw=none},
        ticklabel style = {font=\scriptsize},
        xtick={-1, -0.5, 0, 0.5, 1},
        ytick={0, 0.25, 0.5, 0.75, 1},
        ztick={0, 0.5, 1, 1.5},
        zmin=0,
        zmax=1.75,
        point meta max=100,
        point meta min=1,
        width=\linewidth,
        height=0.74\linewidth,
    ]%
    \addplot3 [surf, ultra thin, line width=0.1pt] table[x=re, y=im, z=norm, col sep=comma] {#1};%
\end{axis}%
\end{tikzpicture}%
}%
\newcommand{\gradPlot}[2]{%
\begin{tikzpicture}[tight background]%
\begin{axis}[%
        colormap/viridis,
        xlabel=Re($\quaternion{z}$),
        ylabel=$\lVert\text{Im}(\quaternion{z})\rVert$,
        zlabel=$\lVert \quaternion{f}(\quaternion{z})^{\prime} \rVert$,
        xlabel style = {sloped, anchor=center},
        ylabel style = {sloped, anchor=center},
        ylabel shift = 3pt,
        zlabel shift = -3pt,
        grid=major,
        axis lines*=left,
        view={-45}{10},
        tick style={draw=none},
        ticklabel style = {font=\scriptsize},
        xtick={-1.5, -1, -0.5, 0, 0.5, 1, 1.5},
        ytick={0, 0.25, 0.5, 0.75, 1, 1.25, 1.5},
        ztick=#2,
        width=\linewidth,
        height=0.74\linewidth,
    ]%
    \addplot3 [mesh, samples=10, domain=-1.5:1.5, y domain=0:1.5, red] {x*0 + y*0};
    \addplot3 [surf, ultra thin, line width=0.1pt] table[x=re, y=im, z=norm, col sep=comma] {#1};%
\end{axis}%
\end{tikzpicture}%
}%
\newcommand{\gradPlotTwo}[3]{%
\begin{tikzpicture}[tight background]%
\begin{axis}[%
        colormap/viridis,
        xlabel=Re($\quaternion{z}$),
        ylabel=$\lVert\text{Im}(\quaternion{z})\rVert$,
        zlabel=$\lVert \quaternion{f}(\quaternion{z})^{\prime} \rVert$,
        xlabel style = {sloped, anchor=center},
        ylabel style = {sloped, anchor=center},
        ylabel shift = 3pt,
        zlabel shift = -3pt,
        grid=major,
        axis lines*=left,
        view={-45}{10},
        tick style={draw=none},
        ticklabel style = {font=\scriptsize},
        xtick={-1.5, -1, -0.5, 0, 0.5, 1, 1.5},
        ytick={0, 0.25, 0.5, 0.75, 1, 1.25, 1.5},
        ztick=#3,
        zmax=1.2,
        width=\linewidth,
        height=0.74\linewidth,
    ]%
    \addplot3 [mesh, samples=10, domain=-1.5:1.5, y domain=0:1.5, red] {x*0 + y*0};
    \addplot3 [surf, ultra thin, line width=0.1pt] table[x=re, y=im, z=norm, col sep=comma] {#1};%
\end{axis}%
\end{tikzpicture}%
}%
\newcommand{\accPlotLine}[3][solid]{
	\addplot[color=#2, draw opacity=0.2, name path=A, forget plot] table[col sep=comma, y=#3_min, x expr=\coordindex] {\csvdata};%
	\addplot[color=#2, draw opacity=0.2, name path=B, forget plot] table[col sep=comma, y=#3_max, x expr=\coordindex] {\csvdata};%
	\addplot[color=#2, opacity=0.2, forget plot] fill between[of=A and B];%
	\addplot[color=#2, #1, line width=1.5pt] table[col sep=comma, y=#3_mean, x expr=\coordindex] {\csvdata};%
}%
\newcommand{\accPlot}[5]{%
    \begin{tikzpicture}[tight background]%
    \pgfplotstableread[col sep=comma]{#1}\csvdata%
	\begin{axis}[%
        xlabel=Epoch,
        ylabel=Accuracy in \%,
        grid=major,
        tick align=outside,
        ticklabel style = {font=\footnotesize},
        xtick pos=left,
        ytick pos=left,
        ymin=#2,
        ymax=#3,
        xmin=#4,
        xmax=#5,
        width=\textwidth,
        height=0.618\textwidth,
        legend pos=south east,
        legend columns=2,
        legend style={font=\footnotesize},
        legend cell align={left},
	]%
        \accPlotLine{matplotlib_blue}{Cardioid_2}%
        \addlegendentry{Cardioid};%
        \accPlotLine{matplotlib_orange}{MagnitudeTanh}%
        \addlegendentry{MagnitudeTanh};%
        \accPlotLine{matplotlib_green}{Norm}%
        \addlegendentry{Norm};%
        \accPlotLine{matplotlib_red}{PhaseTanh_halfAngle}%
        \addlegendentry{PhaseTanh};%
        \accPlotLine{matplotlib_purple}{PhaseTanhshrink_halfAngle}%
        \addlegendentry{PhaseTanhshrink};%
        \accPlotLine{matplotlib_brown}{NormalizedPhaseTanh_psi}%
        \addlegendentry{ScaledPhaseTanh};%
        \accPlotLine{matplotlib_pink}{NormalizedPhaseTanhshrink_psi}%
        \addlegendentry{ScaledPhaseTanhshrink};%
        \accPlotLine{matplotlib_grey}{PhaseSin_psi_2}%
        \addlegendentry{PhaseSin};%
        \accPlotLine{matplotlib_yellow}{NormPhaseSin_psi}%
        \addlegendentry{ScaledPhaseSin};%
        \addlegendimage{empty legend}\addlegendentry{};%
        \accPlotLine[dashed]{matplotlib_blue}{ReLU}%
        \addlegendentry{Split-ReLU};%
        \accPlotLine[dashed]{matplotlib_orange}{Tanh}%
        \addlegendentry{Split-Tanh};%
	\end{axis}%
    \end{tikzpicture}%
}%
\pgfmathsetmacro{\PI}{3.141592654}%
\pgfplotsset{
    /pgfplots/colormap={hsv_mod}{
        rgb255=(255,0,0) 
        rgb255=(255,255,0) 
        rgb255=(0,255,0)
        rgb255=(0,255,255) 
        rgb255=(0,0,255) 
        rgb255=(255,0,255) 
        rgb255=(255,0,100)
    } 
}%
\pgfplotsset{
    /pgfplots/colormap={gist_rainbow}{
        rgb255=(255,0,40) rgb255=(255,93,0) rgb255=(255,234,0)
        rgb255=(140,255,0) rgb255=(0,255,0) rgb255=(0,255,139)
        rgb255=(0,235,255) rgb255=(0,94,255) rgb255=(41,0,255)
        rgb255=(182,0,255) rgb255=(255,0,191)
    } 
}%
\newcommand{\phasePlot}[1]{
    \begin{tikzpicture}[tight background]%
    \begin{axis}[%
            colormap name=gist_rainbow,
            xlabel=Re($\quaternion{z}$),
            ylabel=$\lVert\text{Im}(\quaternion{z})\rVert$,
            tick align=outside,
            tick pos=left,
            ticklabel style = {font=\scriptsize},
            width=\linewidth - 28pt,
            height=0.6\linewidth,
            colorbar,
            colorbar style={
                ytick={0, \PI/2, \PI},
                yticklabels={0, $\frac{\pi}{2}$, $\pi$},
                yticklabel style={inner sep=0pt},
                ylabel=$\psi(\quaternion{z})$,
                tick align=outside,
                tick pos = right,
                at={(1.03,1)},
            },
            colorbar/width=7pt,
            point meta min=0,
            point meta max=\PI,
            mesh/ordering=y varies,
            point meta=explicit,
            enlargelimits=false,
            axis line style = thick,
        ]%
        \addplot [matrix plot*] table[x=re, y=im, meta=phase, col sep=comma] {#1};%
    \end{axis}
\end{tikzpicture}%
}
\section{Introduction}

Activation functions are crucial for the success of \acp{NN} and deep learning. Not only do they enable to learn non-linear relationships,  but the choice of activation function can significantly affect the performance of the \ac{NN} \cite{dubey_activation_2022}. The usage of the ReLU activation function for example was crucial for the success of AlexNet \cite{AlexNet}, one of the breakthroughs of deep learning, as it enabled a way larger \ac{NN} model by significantly reducing the required training time.

With the transition of \ac{NN} to number systems other than real numbers, the question of appropriate activation functions becomes a very relevant topic. In complex \ac{NN} for example, various approaches exist, however as surveyed by \cite{bassey_survey_2021}, no clear consensus has been achieved yet.
For quaternion valued \ac{NN}, the vast majority of approaches utilize the so called elementwise or split activation, applying a known activation like e.g. ReLU on the real and imaginary parts separately. Although being a straight forward approach, this imposes potential weaknesses in the architecture:
Elementwise activations do not respect the $\mathbb{H}$-space, instead this strategy can be interpreted as a mapping from $\mathbb{H} \rightarrow \mathbb{R}^4$ and then $\mathbb{R}^4 \rightarrow \mathbb{H}$. Also, applying e.g. a split ReLU on the quaternion components is hardly interpretable, especially from a geometric viewpoint. Furthermore, \ac{QNN} thrive from their ability to enhance the interrelationship of the respective real and imaginary parts, which is neglected by split activations.

Contrary, \cite{greenblatt_introducing_2018,che_ujang_quaternion-valued_2011,yili_xia_quaternion-valued_2015}, already utilize quaternion properties in their activations, but they have the downside of requiring special learning rules or containing singularities. Alternatively, \cite{valle_quaternion-valued_2020} utilizes normalization of the quaternions as activation, whereas \cite{kinugawa_proposal_2016} proposes a so-called isotropic activation function used with pure quaternions.
Nevertheless, the question of the effect of different activation functions on the performance of \acp{QNN} and especially potential advantages of activation functions utilizing quaternion properties remains an open topic worth further investigation.

In this paper, we aim to evaluate, how novel activation functions utilizing quaternion properties perform in comparison to established, elementwise / split activation functions. Specifically, we develop criteria for activation functions in quaternion space, on which we base our subsequently proposed activation functions.
The core idea is to utilize the polar representations of quaternions and introduce the non-linearity by modifying the magnitude or phase of the respective quaternions. Precisely, we propose various quaternion activation functions and compare them with the ReLU and Tanh activation function applied elementwise. We further generalize \cite{kinugawa_proposal_2016} and include the normalization as used in \cite{valle_quaternion-valued_2020}.

Our major contributions are the following:

\begin{itemize}
    \item We propose quaternion activation functions utilizing the magnitude and phase of the quaternions polar representation as an alternative to the commonly used split activation functions. These activation functions follow defined criteria for activations within \ac{QNN} and are designed to utilize all quaternion components for the calculation of the respective output quaternion components.
    \item We provide insights about the gradients of the proposed activation functions based on the GHR calculus \cite{xu_enabling_2015}, where especially the activation functions modifying the phase are characterized by a very high sensitivity over the entire possible input range.
    \item We investigate the effect of using two different angle definitions: the angle between the real part and the imaginary vector as well as the angle of rotation encoded in a unit quaternion within our proposed activation functions on the prediction performance.
    \item We experimentally evaluate the effectiveness of the proposed quaternion activation functions in comparison to the typically used elementwise / split activations using VGG-style \ac{NN} architectures with varying depth and parameter counts, the CIFAR-10 and the SVHN Dataset. 
    In these experiments, in particular our proposed activation functions affecting consistently outperform the commonly used split activation approach.
\end{itemize}

The proposed activation functions can be incorporated in arbitrary \ac{QNN} architectures and used in gradient based training setups. Thus, they have the potential to improve already existing applications of quaternion valued \acp{NN} by replacing the used activation function.

This paper is structured as follows: Section \ref{sec:related_work} presents the related work, and Section \ref{sec:fundamentals} the required quaternion preliminaries. This is followed by our proposed quaternion activation functions in Section \ref{sec:quaternion_activations}, their derivatives in Section \ref{sec:derivatives} as well as the experimental evaluation in Section \ref{sec:experiments}. Section \ref{sec:conclusion} concludes the paper.
\section{Related Work}
\label{sec:related_work}

In this section, we present the related work, which can be divided into papers targeting regular activations / activations for real valued \ac{NN}, activation functions for complex valued \ac{NN} and finally activations used in \ac{QNN}.

\subsection{Regular activations}

In initial work on machine learning and neural networks, often the step function was the activation function of choice. However, it was not suitable for gradient descent based optimization as it is not continuously differentiable. 
Eventually, the step function was replaced by the sigmoid function as e.g. in the well known backpropagation paper \cite{rumelhart_learning_1986} or the tangens hyperbolicus (Tanh). However, for deeper models, these activation functions are susceptible to vanishing gradients \cite{bengio_learning_1994} due to gradient magnitudes $<1$ and saturation effects. 
This is addressed by the ReLU activation function \cite{Glorot2011DeepSR, Maas2013RectifierNI}
which at the same time also yields sparsity, an often beneficial and desired property.

Based on the success of ReLU, many adaptions and improvements were developed. The first to name  here is the LeakyReLU \cite{Maas2013RectifierNI}. ReLU has the disadvantage that the gradient for non-activated neurons is zero which can yield dead neurons. This is prevented by a small slope for $x < 0$ which yields gradients unequal to zero for the entire input range.
In \cite{ELU}, the Exponential Linear Unit (ELU) is introduced. ELU is intended to improve learning of \ac{NN} by pushing the mean activation closer to zero, similar to batch normalization.
The authors of \cite{he_delving_2015} introduce PReLU, a Parametric Rectified Linear Unit, which is similar to LeakyReLU, but the slope is controlled by a learnable parameter. Countless further activation functions were proposed, be it trainable or non-trainable ones. An overview can be found in e.g. \cite{apicella_survey_2021} or \cite{dubey_activation_2022}. Based on the listing above, we can see the constant development within the field of activation functions, highlighting its relevance to further improve \ac{NN} architectures. 
Consequently, designing appropriate activation functions is the quaternion space $\mathbb{H}$ is of similar importance, providing the motivation for our work.

\subsection{Complex activations}

With the advent of complex valued neural networks, simultaneously the question of suitable activation functions in the complex plane became relevant, and various variants were proposed. However, as stated by \cite{bassey_survey_2021}, no favorite activation function has yet emerged. A straight forward and simple solution which is among the most used is the element-wise application of known and established activations from the conventional \ac{NN} as in \cite{hirose_generalization_2012, benvenuto_complex_1992, kinouchi_memorization_1996, hayakawa_applying_2018, popa_complex-valued_2017, peker_novel_2016, ishizuka_modeling_2018}.
A different approach uses the polar representation of complex numbers, where conventional activation functions are applied to the magnitude and/or phase of the respective complex number.
This approach is pursued by \cite{hirose_continuous_1992, georgiou_complex_1992, hirose_generalization_2012, hayakawa_applying_2018, mizote_optical_2013, al-nuaimi_enhancing_2012}. 

Besides these activation techniques, further less commonly used ones exist: The first one is the complex cardioid, as proposed in \cite{virtue_better_2017}. Here, the phase of the input is used to 
control the magnitude of the output complex number. When the imaginary part is zero, it resembles the ReLU activation. Further, \cite{scardapane_complex-valued_2020} proposes split kernel activation and complex kernel activation functions. A different approach is pursued by \cite{jankowski_complex-valued_1996}, where the activation input is mapped to $K$ equidistant values on the unit circle in the complex plane to achieve discrete neuron outputs to be used in a Hopfield network. This strategy can also be found in \cite{kobayashi_o2_2019-2, kobayashi_decomposition_2018}.

Finally, \cite{kim_fully_2002} proposes a set of fully complex activation functions. However, they can contain singularities or discontinuities, which imposes the requirement of careful handling during training. A fully complex Tanh is also stated in \cite{mandic_complex_2009}. It is evident that leaving the real number space as the mathematical system for modelling \ac{NN} architectures introduces further challenges in suitable activation functions and emphasizes the relevance of research in that direction.

\subsection{Quaternion activations}
\label{subsec:quatActivations}

Transitioning from the complex \ac{NN} to the hypercomplex ones, specifically the \ac{QNN}, the question of appropriate activation functions remained. 

By far the most popular variant is the element-wise application of known activation functions on the real part and the three imaginary parts of a quaternion, the so called split activation function, like the
split Tanh \cite{Parcollet2019, xu_optimization_2016}, 
split HardTanh \cite{parcollet_quaternion_2019}, 
split Sigmoid \cite{zhang_quaternion_2019},
split ReLU \cite{chen_quaternion_2023,comminiello_quaternion_2019,comminiello_quaternion_2019, Zhu_2018_ECCV, tay_lightweight_2019, muppidi_speech_2021, gai_reduced_2022},
split LeakyReLU \cite{grassucci_quaternion-valued_2021} or
a mixture of multiple split activations \cite{razavi-far_quaternion_2022}.
Similarly, elementwise applications of other functions such as nonmonotonic piecewise nonlinear activation functions \cite{tan_multistability_2019}, discontinuous activation functions \cite{wei_fixed-time_2023} or trainable Bessel activation functions \cite{vieira_quaternionic_2023} are used.

Similarly, many further variants can be found.
In \cite{greenblatt_introducing_2018}, the quaternions get decomposed in $e^{i \phi}e^{j \theta}e^{k \psi}$, normed and quantized. 
This, however, requires special learning rules as is it not continuous and is hence not generally applicable to standard \ac{QNN} architectures.
Alternatively, \cite{valle_quaternion-valued_2020} uses normalization to unit quaternions as the activation function.
Similarly, \cite{che_ujang_quaternion-valued_2011} and \cite{yili_xia_quaternion-valued_2015} use a fully quaternion version of the Tanh. Unfortunately, this has singularities as claimed by \cite{Parcollet2019} what hinders training and limits the widespread use.
Also utilizing quaternion properties, \cite{kinugawa_proposal_2016} proposes an isotropic activation to be used in combination with pure quaternions. Contrary, \cite{qin_fast_2022-2} designed custom quaternion product units which are non-linear itself, hence don't require a further non-linearity.

However, all approaches lack a systematic design of  activation functions based on carefully defined criteria, incorporating general properties valid for arbitrary number spaces as well as properties specifically for the quaternion space. Further, a detailed experimental comparison of different activation approaches has not been conducted.
Furthermore, we propose activation functions based on modifying the magnitude or phase, utilizing the polar representation of quaternions. To the best of our knowledge, no work exist utilizing the quaternion phase within the activation function and no work providing such a comparison of the different activation approaches.

\section{Preliminaries}
\label{sec:fundamentals}

In  the following, we introduce the required preliminaries, namely the quaternion algebra and the quaternion neural networks.

\subsection{Quaternion Algebra}

Hamilton discovered quaternions in 1843 \cite{hamilton_ii_1844} during his endeavor to expand complex numbers to three dimensions. This required three imaginary units $\imagI, \imagJ, \imagK$ with the properties
\begin{equation}
\begin{aligned}
    \imagI^2 = \imagJ^2 = \imagK^2 = \imagI\imagJ\imagK = -1 \\
    \imagI\imagJ = +\imagK,~~ \imagJ\imagK = +\imagI,~~ \imagK\imagI = +\imagJ \\
    \imagJ\imagI = -\imagK,~~ \imagK\imagJ = -\imagI,~~ \imagI\imagK = -\imagJ .
\end{aligned}
\end{equation}
This allows to formulate a quaternion $\quaternion{q}$ as
\begin{equation}
    \quaternion{q} = \quaternionComponents{q} = \quatCompR{q} + \mathbf{q}
    \label{equ:quat_standard_form}
\end{equation}
yielding the set of quaternions $\mathbb{H} = \{\quaternion{q}: \quaternion{q} = \quaternionComponents{q},~ \quatCompR{q}, \quatCompI{q}, \quatCompJ{q}, \quatCompK{q} \in \mathbb{R}\}$
where  $\quatCompR{q}$ is the real part and $\mathbf{q}$ the imaginary vector.
As in the complex numbers, a quaternions conjugate is defined:
\begin{equation}
    \quatConj{q} = \quaternionConjComponents{q} = \quatCompR{q} - \mathbf{q} .
\end{equation}

Two quaternions $\quaternion{x}, \quaternion{y}$ can be added element-wise following
\begin{equation}
    \quaternion{x} + \quaternion{y} = (\quatCompR{x} + \quatCompR{y}) + (\quatCompI{x} + \quatCompI{y})i + (\quatCompJ{x} + \quatCompJ{y})j + (\quatCompK{x} + \quatCompK{y})k
\end{equation}
and multiplied following 
\begin{equation}
\begin{aligned}
    x \otimes y &= \quatCompR{x}\quatCompR{y} - \mathbf{x} \cdot \mathbf{y} +  x_0\mathbf{y}+y_0\mathbf{x} + \mathbf{x} \times \mathbf{y} \\
	&= ( \quatCompR{x} \quatCompR{y} - \quatCompI{x} \quatCompI{y} - \quatCompJ{x} \quatCompJ{y} - \quatCompK{x} \quatCompK{y} )   \\
	&+ ( \quatCompR{x} \quatCompI{y} + \quatCompI{x} \quatCompR{y} + \quatCompJ{x} \quatCompK{y} - \quatCompK{x} \quatCompJ{y} ) \imagI \\
	&+ ( \quatCompR{x} \quatCompJ{y} - \quatCompI{x} \quatCompK{y} + \quatCompJ{x} \quatCompR{y} + \quatCompK{x} \quatCompI{y} ) \imagJ \\
	&+ ( \quatCompR{x} \quatCompK{y} + \quatCompI{x} \quatCompJ{y} - \quatCompJ{x} \quatCompI{y} + \quatCompK{x} \quatCompR{y} ) \imagK .
\end{aligned}
\label{equ:hamilton_product}
\end{equation}
Quaternions $\quaternion{q}$ satisfying $\left\lVert \quaternion{q} \right\rVert = \sqrt{\quaternion{q}\quatConj{q}} = \sqrt{\quatCompR{q}^2 + \quatCompI{q}^2 + \quatCompJ{q}^2 + \quatCompK{q}^2} = 1$ are usually referred to as unit quaternions and quaternions $\quaternion{q}$ where $\quatCompR{q} = 0$ are referred to as pure quaternions.
These unit quaternions can be used to encode a rotation following
\begin{equation}
    \quaternion{q} = \cos{\left( \frac{\theta}{2} \right)} + \mathbf{n} \sin{\left( \frac{\theta}{2} \right)} ,
    \label{equ:quaternion_rotation_angle}
\end{equation}
where $\theta$ is the angle of rotation and $\mathbf{n}$ is the axis of rotation satisfying $\lVert \mathbf{n} \rVert = 1$.
Alternatively, this is sometimes also denoted as 
\begin{equation}
    \quaternion{q} = \cos{\left( \psi \right)} + \mathbf{n} \sin{\left( \psi \right)}.
    \label{equ:quaternion_phase}
\end{equation}
Using this notation, we can obtain the angle $\psi$ or the phase of the quaternion using \cite{dantam_robust_2021}
\begin{equation}
    \psi = \atan2(\lVert \mathbf{q} \rVert, \quatCompR{q}),
    \label{equ:angle_atan_representation}
\end{equation}
while 
\begin{equation}
    \sin{(\psi)} = \frac{\lVert \mathbf{q} \rVert}{\lVert \quaternion{q} \rVert}
\label{equ:angle_sin_representation}
\end{equation}
and
\begin{equation}
    \cos{(\psi)} = \frac{\quatCompR{q}}{\lVert \quaternion{q} \rVert}.
    \label{equ:angle_psi-cos}
\end{equation}
Note the two different angle notations $\theta$ and $\psi$ in Equations \eqref{equ:quaternion_rotation_angle} and \eqref{equ:quaternion_phase} which are required later on.
The angle $\theta$ represents the encoded rotation in a unit quaternion, whereas $\psi$ is the general angle between the real part and the imaginary vector. Both can be a valid choice to be used in an activation function, however as the default we use $\psi$ as this is the commonly used phase definition, similar to that in the complex numbers.

To convert the quaternion form of Equation \eqref{equ:quat_standard_form} to the one in Equation \eqref{equ:quaternion_phase} when dealing with non-unit quaternions, we can use the following relation:
\begin{equation}
\begin{aligned}
    \quaternion{q} &= \quaternionComponents{q} \\
    &= \lVert \quaternion{q} \rVert \frac{\quaternion{q}}{\lVert \quaternion{q} \rVert} 
     = \lVert \quaternion{q} \rVert \frac{\quaternionComponents{q}}{\lVert \quaternion{q} \rVert} \\
    &= \lVert \quaternion{q} \rVert \left( 
         \frac{\quatCompR{q}}{\lVert \quaternion{q} \rVert} + 
         \frac{\mathbf{q}}{\lVert \quaternion{q} \rVert}
    \right)
    = \lVert \quaternion{q} \rVert \left( 
         \frac{\quatCompR{q}}{\lVert \quaternion{q} \rVert} + 
         \frac{\mathbf{q}}{\lVert \quaternion{q} \rVert} \frac{\lVert \mathbf{q} \rVert}{\lVert \mathbf{q} \rVert}
    \right) \\
    &= \lVert \quaternion{q} \rVert \left( 
         \frac{\quatCompR{q}}{\lVert \quaternion{q} \rVert} + 
         \frac{\mathbf{q}}{\lVert \mathbf{q} \rVert} \frac{\lVert \mathbf{q} \rVert}{\lVert \quaternion{q} \rVert}
    \right)
    = \lVert \quaternion{q} \rVert \left( 
         \cos{(\psi)} + 
         \frac{\mathbf{q}}{\lVert \mathbf{q} \rVert}  \sin{(\psi)}
    \right) \\
    &= \lVert \quaternion{q} \rVert \left( 
         \cos{(\psi)} + 
         \frac{\mathbf{q}}{\lVert \quaternion{q} \rVert \sin{(\psi)}}  \sin{(\psi)}
    \right) . \\
\end{aligned}
\end{equation}

Hence, we can obtain the rotation axis $\mathbf{n}$ using either $\mathbf{n} = \frac{\mathbf{q}}{\lVert \mathbf{q} \rVert}$ or $\mathbf{n} = \frac{\mathbf{q}}{\lVert \quaternion{q} \rVert \sin{(\psi)}}$.

\subsection{Quaternion Neural Networks}

We proceed by introducing the \ac{QNN} layer used within this work, based on the above introduced quaternion algebra.

\subsubsection{Quaternion Linear Layer}

We can formulate the forward phase of a regular feed-forward \ac{QNN} layer $(l)$ with $n$ inputs and $m$ outputs as follows:

The output $\quaternion{a}_i$ of such a quaternion linear layer $(l)$ is calculated following

\begin{equation}
\begin{aligned}
    \quaternion{a}_i^{(l)} &= \sigma(\quaternion{z}_i^{(l)}),~~ 
    \quaternion{z}_i^{(l)} 
    = \sum_{j=1}^{n} \quaternion{w}_{i,j}^{(l)} \quaternion{a}_j^{(l-1)} + \quaternion{b}_i^{(l)}
\end{aligned}
\end{equation}
where $i \in \{1, \dots, m\}$, $j \in \{1, \dots, n\}$ and $\quaternion{w}, \quaternion{y}, \quaternion{b}, \quaternion{a}, \quaternion{z} \in \mathbb{H}$. $\sigma(\cdot)$ denotes an arbitrary activation function. 

The matrix-vector calculation of a quaternion linear layer consequently is

\begin{equation}
\begin{aligned}
    \quatVec{a}^{(l)} &= \sigma(\quatVec{z}^{(l)}), ~~ 
    \quatVec{z}^{(l)} 
    = \quatVec{W}^{(l)} \quatVec{a}^{(l-1)} + \quatVec{b}^{(l)}
\end{aligned}
\label{equ:layerMatrixVector}
\end{equation}
where $\quatVec{W} \in \mathbb{H}^{m \times n}$ and $\quatVec{b} \in \mathbb{H}^{m}$.

\subsubsection{2D Quaternion Convolution Layer}

In order to describe the 2D quaternion convolution, consider a two-dimensional quaternion input $\quaternion{F} \in \mathbb{H}^{m \times n}$. For the convolution with a kernel $\quaternion{w} \in \mathbb{H}$, we obtain the output $\quaternion{G}(x, y)$ by applying
\begin{equation}
    \quaternion{G}(x, y) = \quaternion{F}(x, y) \ast \quaternion{w} 
    = \sum_{\delta x=-k_i}^{k_i}\sum_{\delta y=-k_j}^{k_j} \quaternion{F}(x + \delta x, y + \delta y) \quaternion{w}(\delta x, \delta y)
\end{equation}
where $-k_j \leq \delta y \leq k_j$ and $-k_j \leq \delta y \leq k_j$ are the kernel elements.

Considering $C_{in}$ input channel and $C_{out}$ output channel, the output of the layer is calculated using
\begin{equation}
    \quaternion{O}(C_{out}, x, y) = \quaternion{b}(C_{out}) + \sum_{c = 0}^{C_{in} - 1} \quaternion{G}_c (x, y)
\end{equation}
where $\quaternion{b} \in \mathbb{H}^{C_{out}}$ is the layer's bias and $\quaternion{G}_c$ utilizes the kernel $\quaternion{w}$ corresponding to the respective input channel $c$.

For ease of notation, we waive on using stride and dilation.

\subsubsection{2d Quaternion Average Pooling}

Assume a pooling kernel of width $w$ and height $h$ striding over a quaternion valued, two-dimensional input, extracting patches $\quaternion{p}_i \in \mathbb{H}^{w \times h}$. Then the pooled output $\quaternion{o} \in \mathbb{H}$ is calculated using
\begin{equation}
    \quaternion{o} = f(\quaternion{p}_i) = \frac{1}{wh} \sum_{i=0}^{wh -1} \quaternion{x_i}
\label{equ:quat_avg_pool}
\end{equation}
where $\quatVec{x} = [\quaternion{x_0}, \quaternion{x_1}, \dots, \quaternion{x_{wh - 1}}]$ is the vector representation of the extracted patch $\quaternion{p}_i$.

\section{Quaternion Activation Functions}
\label{sec:quaternion_activations}

In this section, we first define criteria to be fulfilled by activation functions in quaternion space, then we follow up with our used quaternion visualization methodology, and finally we propose our novel quaternion activation functions.

\subsection{Criteria for Quaternion Activation Functions}
\label{subsec:criterias}

Before we start to introduce and discuss the quaternion activation functions, first we want to define criteria which we aim for with our proposed activation functions, to obtain the best possible performance for \acp{QNN}.
To this end, we define general criteria for the design of activation functions where we derive general criteria valid for arbitrary number spaces and criteria specifically tailored for quaternion spaces.
Based on previous experiences on \ac{NN} training and representation capabilities, we state a first set of criteria as follows: 
\begin{enumerate}
    \item Sufficient non-linearity. To enable the learning of non-linear relationships for \acp{QNN}, our proposed activation functions need to introduce sufficient non-linearity into the model. \label{enum:non-linearity}
    \item Suitability for gradient based optimization. We aim for activation functions that are fully differentiable, such that they can be trained with established methods and are generally applicable, different to e.g. \cite{greenblatt_introducing_2018}. \label{enum:optim}
    \item Proper gradient flow with maximized sensitivity. We aim to propose activation functions which yield gradients for a maximum-sized input range, independent of sign or numerical value. \label{enum:gradient-flow}
\end{enumerate}

The second set of criteria is specifically tailored to the quaternion properties of \acp{QNN}. 
Specifically, they are:

\begin{enumerate}[resume]
    \item Utilization of all quaternion components of the input quaternion during calculation of the respective output components: Assume a quaternion activation function $\quaternion{f}$ such that $\quaternion{a} = \quaternion{f}(\quaternion{z})$. Then, we require $\quatCompR{a} = f(\quatCompR{z},\quatCompI{z},\quatCompJ{z},\quatCompK{z})$,  $\quatCompI{a} = f(\quatCompR{z},\quatCompI{z},\quatCompJ{z},\quatCompK{z})$,  $\quatCompJ{a} = f(\quatCompR{z},\quatCompI{z},\quatCompJ{z},\quatCompK{z})$ and  $\quatCompK{a} = f(\quatCompR{z},\quatCompI{z},\quatCompJ{z},\quatCompK{z})$. Thereby, we suggest to incorporate the well known mechanism as in the Hamilton product within the quaternion convolution \cite{parcollet_quaternion_2019, gaudet_deep_2018, poppelbaum_time_2024} into the activation function, namely explicitly linking the respective components during calculation. \label{enum:quat_components}
    \item Preservation of the ratios between the respective quaternion components: This is a category for activation functions with the intent to modify the quaternion while keeping the ratio between the quaternion components unchanged. Consequently, such designed activation functions amplify or dampen the quaternions with their incorporated extracted features.
    \label{enum:preservation_ratios}
    \item Preservation of the quaternion magnitude: We aim for activation functions $\quaternion{a} = \quaternion{f}(\quaternion{z})$ that leave the quaternion magnitude unchanged but instead incorporate non-linearity by modifying the respective components $\quatCompR{z},\quatCompI{z},\quatCompJ{z}$ and $\quatCompK{z}$ such that $\lVert \quaternion{a} \rVert = \lVert \quaternion{z} \rVert$ This criterion leverages the multidimensionality of the quaternion and is automatically unit preserving. Furthermore, this design prevents saturation effects.
    \label{enum:preservation_magnitude}
    \item Interpretability: We aim for activation functions allowing to interpret how the non-linearity was introduced into the quaternion input space and how it modifies the respective quaternion. In particular, we seek to understand the quaternion modification from a geometrical viewpoint, not just from raw calculation.\label{enum:interpretability}
    \item Ability to be used within arbitrary \ac{QNN} architectures: We aim for activation functions that can be incorporated in arbitrary model architectures and styles, like MLP-based, CNN-based, RNN-based or any other form that requires non-linearities within the layers. This allows for widespread usage and incorporation in already existing \ac{QNN} applications. \label{enum:arbitrary_architectures}
\end{enumerate}

\begin{remark}
We note, that Criteria \ref{enum:preservation_ratios} and \ref{enum:preservation_magnitude} cannot be fulfilled at the same time. Instead, they define two different categories of quaternion activation function behavior.
\end{remark}

\begin{remark}
When we apply the defined criteria to the commonly used split activations, it is easy to see, that they do not fulfill criterion \ref{enum:quat_components} as we have $\quatCompR{a} = f(\quatCompR{z})$, $\quatCompI{a} = f(\quatCompI{z})$, $\quatCompJ{a} = f(\quatCompJ{z})$ and $\quatCompK{a} = f(\quatCompK{z})$. Further, neither \ref{enum:preservation_ratios} nor \ref{enum:preservation_magnitude} is fulfilled, as the respective components are modified using algebraic operation designed for vectors in $\mathbb{R}$ and not in $\mathbb{H}$. These modifications and their influence on the whole quaternion, especially on its magnitude and phase, is hardly, if at all, interpretable. Therefore, we claim that criterion \ref{enum:interpretability} is not fulfilled, too.
\end{remark}

\subsection{Visualizing the Quaternion Activation Functions}

In order to analyze criterion \ref{enum:preservation_magnitude} and the gradients of the respective activation functions according to criterion \ref{enum:gradient-flow}, we require for a suitable visualization of the proposed quaternion activations and their gradients. However, due to their multidimensional nature with four dimensions, quaternions are inherently difficult to visualize. Nevertheless, a visualization is very helpful to understand the behavior of activation functions.

\begin{figure*}[htb]
    \centering%
    \begin{subfigure}[c]{0.5\textwidth}%
        \centering%
 		\includegraphics{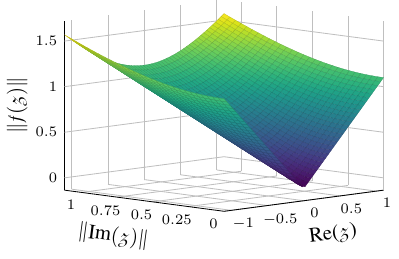}
        \caption{Magnitude of the quaternion meshgrid}%
    \end{subfigure}%
    \hfill%
    \begin{subfigure}[c]{0.5\textwidth}%
        \centering%
		\includegraphics{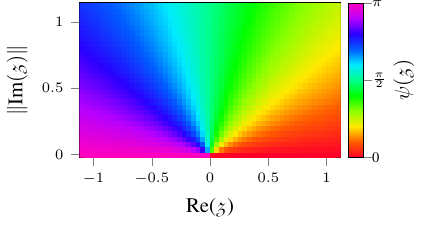}
        \caption{Phase of the quaternion meshgrid}%
        \label{fig:meshgrid_init-phase}
    \end{subfigure}%
    \caption{Visualization of the quaternion meshgrid used as an input for the visualization of the quaternion activation functions.}%
    \label{fig:meshgrid_init}
\end{figure*}

Thus, a dimensionality reduction is required to allow for a suitable visualization of the four dimensional quaternions after activation or activation gradients.
To obtain this reduction, we opt for the following approach:
Assume a function $\quaternion{a} = \quaternion{f}(\quaternion{z}),~\quaternion{a}, \quaternion{z} \in \mathbb{H}$.
Then, as in the complex numbers, we create a mesh-grid in x- and y-direction. For the x-direction, we use the real part $\text{Re}(\quaternion{z}) = \quatCompR{z}$ of $\quaternion{z}$ and for the y-direction we utilize the norm of the imaginary parts $\lVert \text{Im}(\quaternion{z}) \rVert = \sqrt{\quatCompI{z}^2 + \quatCompK{z}^2 + \quatCompK{z}^2}$. This reduces the three-dimensional imaginary vector to a scalar, yielding the required dimensionality reduction. 
Further, we want to visualize two properties: the magnitude and the phase. For the magnitude, we select a 3D-plot where $\lVert \quaternion{a} \rVert = \lVert \quaternion{f}(\quaternion{z}) \rVert$, hence the magnitude of the output of $\quaternion{f}()$, is plotted on the z-axis. 
The phase-plot is selected as 2D where the color indicates the respective phase of the functions output $\quaternion{a}$.
An example of this visualization method with just the meshgrid without any activation applied is shown in Figure \ref{fig:meshgrid_init}.

\begin{remark}
    This quaternion meshgrid is not unique as there are endless possibilities to obtain a certain norm for the imaginary vector. A norm of $\lVert \text{Im} \rVert = 1$ can e.g. be obtained with $\quatCompI{z} = \quatCompJ{z} = \quatCompK{z} = \frac{1}{\sqrt{3}}$ as well as with $\quatCompI{z} = \frac{-1}{\sqrt{14}},~\quatCompJ{z} = \frac{2}{\sqrt{14}},~\quatCompK{z} = \frac{-3}{\sqrt{14}}$. Nevertheless, irrespective of the chosen imaginary components, the overall magnitude is always the same. The same applies for the phase (compare Equations \eqref{equ:angle_atan_representation} - \eqref{equ:angle_psi-cos}).
    Consequently, despite the non uniqueness, this does not impose any restrictions for the chosen visualization approach.
\end{remark}

\subsection{Proposed Quaternion Activation Functions}

In this section, we propose our novel quaternion activation functions. However, to further motivate the relevance of the novel functions, we first recall the by far most commonly used activation approach within \acp{QNN}, namely the split activation approach (compare Subsection \ref{subsec:quatActivations}), and discuss its weaknesses, in particular with regard to the defined criteria. The split activation or element-wise activation is calculated by
\begin{equation}
    \sigma(\quaternion{q}) = \sigma(\quatCompR{q}) + \sigma(\quatCompI{q}) \imagI + \sigma(\quatCompJ{q}) \imagJ + \sigma(\quatCompK{q}) \imagK
\label{equ:split-act}
\end{equation}
where $\sigma(\cdot)$ denotes an activation function in $\mathbb{R}$, e.g. Tanh or ReLU. However, this approach ignores the $\mathbb{H}$-space and treats the quaternions more as $\mathbb{R}^4$. 
Further, one of the great benefits of \acp{QNN} is their ability to utilize the interrelationship of the respective quaternion components due to the Hamilton product, e.g. within the convolution. This is not carried out to the activation function when using the split approach as in Equation \eqref{equ:split-act}, and consequently criterion \ref{enum:quat_components} is not fulfilled.
Also, it is hardly interpretable, especially from a geometric viewpoint. 
Thus, in the following we propose and discuss novel quaternion activation functions which take into account the quaternion properties (see criterion \ref{enum:quat_components}).

Every quaternion $\quaternion{z} = \quaternionComponents{z} = \quatCompR{z} + \mathbf{z}$ can be expressed in polar form as
\begin{equation}
	\quaternion{z} 
	= \lVert \quaternion{z} \rVert (\cos \psi + \mathbf{n} \sin \psi)
	= \lVert \quaternion{z} \rVert e^{\mathbf{n} \psi} .
 \label{equ:quat_polar_form}
\end{equation}
This form allows us to think of quaternions differently as in the regular Cartesian form. 
Specifically, inspired by examples in the complex domain \cite{hirose_continuous_1992, georgiou_complex_1992, hirose_generalization_2012, hayakawa_applying_2018, mizote_optical_2013, al-nuaimi_enhancing_2012}, we advocate for utilizing the magnitude and phase to introduce non-linearity in \acp{QNN}. 
We define two categories of activation functions: functions affecting the magnitude where the phase remains unchanged and functions affecting the phase where the magnitude remains unchanged. For both, we propose variants in the following subsections.

\subsubsection{Activations affecting the Magnitude}

The first category we discuss are activation functions affecting the quaternion magnitude. With these activation functions, we aim to preserve the relation between the respective quaternion components as required by criterion \ref{enum:preservation_ratios}, while still introducing non-linearity. 
When applying a non-linearity on the magnitude, this causes all components $\quatCompR{z}$, $\quatCompI{z}$, $\quatCompJ{z}$ and $\quatCompK{z}$ to be multiplied with the same factor, yielding the preservation. Simultaneously, as all components are required for magnitude calculation, this automatically fulfills criterion \ref{enum:quat_components}. With this strategy, the whole quaternion is either emphasized or damped in comparison to other quaternion inputs.

\begin{figure*}[h]
    \centering
    \begin{subfigure}[c]{0.49\textwidth}
        \centering
		\includegraphics{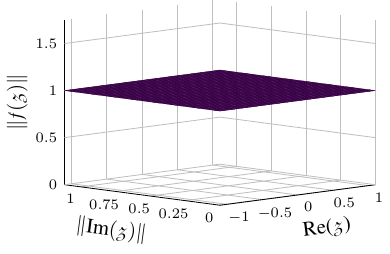}
        \caption{Magnitude}
    \end{subfigure}
    \hfill
    \begin{subfigure}[c]{0.49\textwidth}
        \centering
		\includegraphics{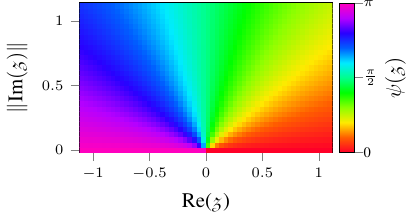}
        \caption{Phase}
    \end{subfigure}
    \caption{Visualization of the Norm activation.}
    \label{fig:Norm}
\end{figure*}

\paragraph{Normalization}
We start with activation through normalization as in \cite{valle_quaternion-valued_2020}. It is calculated following
\begin{equation}
    \phi_{\text{N}} (\quaternion{z}) 
    = 
    \frac{\quaternion{z}}{\lVert \quaternion{z} \rVert}
    =
    \frac{\lVert \quaternion{z} \rVert}{\lVert \quaternion{z} \rVert} e^{\mathbf{n} \psi}
    = 
    \frac{\quaternionComponents{z}}{\quatNorm{z}} .
\end{equation}
The corresponding visualization is shown in Figure \ref{fig:Norm}.
With this strategy, we obtain a mapping to unit quaternions as the output of each layer after activation, ensuring that the inputs for the respective following layers are always in a similar range. This might also be beneficial for learning problems where a unit quaternion output is desired to also force the intermediate results to unit property as well. It is a bounded activation function that emphasizes lower magnitudes to one while it dampens higher magnitudes.
We can now use this activation and extend it since it yields the $e^{\mathbf{n} \psi}$ part of Equation \eqref{equ:quat_polar_form}. Hence, it can serve as a building block for further activations.

\paragraph{MagnitudeTanh}

The previously presented activation by normalization has an equalizing effect. However, this might not always beneficial. Instead, it might be more beneficial for learning problems to dampen or amplifying certain inputs in comparison to others. Consequently, we further want to propose an activation function that obeys this characteristic within the quaternion space. To this end, we opt for the quaternion magnitude and apply the non-linearity to it.

Per definition, the magnitude $\lVert \quaternion{z} \rVert$ of a quaternion $\quaternion{z}$ is always $\geq 0$.
This limits the pool of suitable candidates if established and known activation functions shall be used as building blocks.
Using e.G. ReLU on the magnitude would just end up in an identity function and hence no non-linearity.
Thus, we need activation functions which exhibit a non-linearity for input values > 0. Among the possible candidates, the Tanh is one of the most frequently used activations, which is why we decide to use this as a building block. Furthermore, in contrast to e.g. the Sigmoid, the $\tanh(x) \rightarrow 0$ when $x \rightarrow 0$, enabling the possibility for values to vanish.

Consequently, we propose the MagnitudeTanh which we define as
\begin{equation}
	\phi_{\text{MT}} (\quaternion{z}) 
    = \tanh(\lVert \quaternion{z} \rVert) e^{\mathbf{n} \psi}
    .
\end{equation}
The Cartesian representation of the proposed activation is 
\begin{equation}
\begin{aligned}
	\phi_{\text{MT}} (\quaternion{z}) 
    &= \tanh(\lVert \quaternion{z} \rVert) \frac{\quaternion{z}}{\lVert \quaternion{z} \rVert} \\
	&= \tanh\left(\quatNorm{z}\right) \frac{\quaternionComponents{z}}{\quatNorm{z}} .
\end{aligned}
\end{equation}
Note, that this activation is the generalization of the isotropic activation in \cite{kinugawa_proposal_2016} from pure quaternions to arbitrary quaternions.
As we can see, the norm is part of the function definition, and all quaternion components $\quatCompR{z}$, $\quatCompI{z}$, $\quatCompJ{z}$ and $\quatCompK{z}$ are involved in the calculation of all output components $\quatCompR{a}$, $\quatCompI{a}$, $\quatCompJ{a}$ and $\quatCompK{a}$.
The MagnitudeTanh is a bounded activation function with a maximum magnitude of the respective output quaternion of one. It emphasizes lower magnitudes while dampening higher ones.

\begin{figure*}[h]
    \centering
    \begin{subfigure}[c]{0.49\textwidth}
        \centering
		\includegraphics{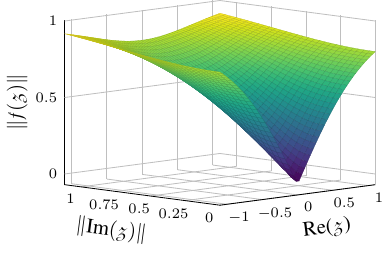}
        \caption{Magnitude}
    \end{subfigure}
    \hfill
    \begin{subfigure}[c]{0.49\textwidth}
        \centering
		\includegraphics{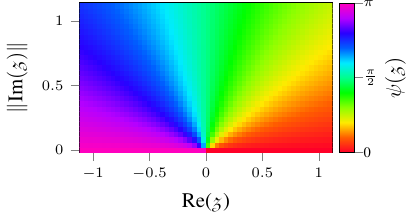}
        \caption{Phase}
    \end{subfigure}
    \caption{Visualization of the MagnitudeTanh.}
    \label{fig:MagTanh}
\end{figure*}

The visualization is shown in Figure \ref{fig:MagTanh}. As intended, just the magnitude of the quaternion is affected and the phase remains unchanged. We also experimented with a MagnitudeTanhshrink, this however did not yield satisfactory results in the experimental evaluation.

\paragraph{Quaternion Cardioid}

It is well known for \ac{NN} operating in the real number space, that ReLU activations improve the training progress compared to the Tanh in a lot of applications due to its unboundedness and improved gradient flow. Consequently, we want to make use of this fact in our quaternion activation functions. The idea is to obtain an unbounded activation function that incorporates the ReLU activation, where we specifically aim for two aspects: reduced saturation effects and improved gradient flow. To this end, we define the Quaternion Cardioid as follows:

\begin{equation}
\begin{aligned}
	\phi_{\text{Cd}} (\quaternion{z}) 
	&= \frac{1}{2} (1 + \cos( \psi )) \quaternion{z} \\
    &\stackrel{\mathmakebox[\widthof{=}]{\eqref{equ:angle_psi-cos}}}{=} \frac{1}{2} (1 + \cos( \arccos(\frac{\quatCompR{z}}{\lVert \quaternion{z} \rVert}) ) \quaternion{z} \\
    &= \frac{1}{2} (1 + \frac{\quatCompR{z}}{\lVert \quaternion{z} \rVert} ) \quaternion{z} 
    .
\end{aligned}
\end{equation}

\begin{figure*}[h]
    \centering
    \begin{subfigure}[c]{0.49\textwidth}
        \centering
		\includegraphics{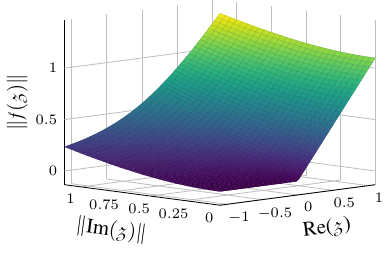}
        \caption{Magnitude}
    \end{subfigure}
    \hfill
    \begin{subfigure}[c]{0.49\textwidth}
        \centering
        \includegraphics{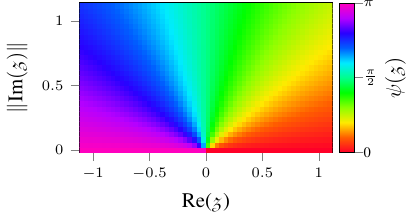}
        \caption{Phase}
    \end{subfigure}
    \caption{Visualization of the QuaternionCardioid.}
    \label{fig:QuaternionCardiodid}
\end{figure*}

Note that the above provides an expansion of the complex cardioid \cite{virtue_better_2017} to the quaternion space where the phase is used to set the magnitude of the activation output.
The corresponding magnitude and phase is shown in Figure \ref{fig:QuaternionCardiodid}. As we can see, contrary to the previous functions and similar to the standard ReLU, the magnitude is not symmetric with the imaginary axis. As in the case of the MagnitudeTanh, the phase remains unchanged. Contrary, the Quaternion Cardioid is an unbounded activation function as long as the real part is not negative and the imaginary vector is zero.
For $\lVert \text{Im}(\quaternion{z}) \rVert = 0$ we end up with the regular ReLU activation function as desired.

\subsubsection{Activations affecting the Phase}

As the second set of quaternion activation functions, we propose activation functions that affect the phase of the quaternion and leave the magnitude unchanged. The idea is to leverage the multi-dimensionality of the quaternion and introduce a non-linearity such that increasing/decreasing the value of one component is at the cost of the other components and vice-versa.
In particular, the change in the phase causes the ratio between the quaternion components to change, without changing the overall magnitude. With this strategy, we achieve an activation where the overall quaternion magnitude remains undamped.
This is particularly useful if e.g. unit property shall be preserved. Note that this activation is non-saturated with respect to the quaternion magnitude as all activations affecting the phase are unbounded ones. Thus, we can explicitly avoid the saturating characteristics of the quaternion cardioid for negative real parts. Note also, that the phase is determined from all quaternion components thereby fulfilling criterion \ref{enum:quat_components}.
Furthermore, this approach yields beneficial properties for the gradient flow, as we will see in the following Section \ref{sec:derivatives}. In the following, we propose the PhaseTanh and the PhaseTanhshrink together with 
a scaled version of both, and finally the PhaseSin and a scaled variant. The PhaseTanh and PhaseTanhshrink use well-known and frequently used activations as building blocks, while the PhaseSin is based on the same principle, but enables a simplification to reduce the complexity of the required calculations.

\paragraph{PhaseTanh and PhaseTanhshrink}
Transferring the idea of the previous activation functions on the phase, we propose the PhaseTanh as
\begin{equation}
\begin{aligned}
	\phi_{\text{PT}} (\quaternion{z}) 
    &= \lVert \quaternion{z} \rVert e^{\mathbf{n} \tanh(\psi)}
	\phi (\quaternion{z}) \\
    &= \lVert \quaternion{z} \rVert \left( 
        \cos(\tanh(\psi)) + \frac{\mathbf{z}}{ \lVert \mathbf{z} \rVert} \sin(\tanh(\psi))
    \right) \\
\end{aligned}
\end{equation}
and the PhaseTanhshrink as
\begin{equation}
\begin{aligned}
	\phi_{\text{PTs}} (\quaternion{z}) 
	&= \lVert \quaternion{z} \rVert e^{\mathbf{n} \tanhshrink(\psi)}
	= \lVert \quaternion{z} \rVert e^{\mathbf{n} [\psi - \tanh(\psi)]} \\
    &= \lVert \quaternion{z} \rVert \left( 
        \cos(\psi -\tanh(\psi)) + \frac{\mathbf{z}}{ \lVert \mathbf{z} \rVert} \sin(\psi - \tanh(\psi))
    \right) . \\
\end{aligned}
\end{equation}

Note that by applying the Tanh as the non-linearity on the phase, the maximum is reached at $\tanh(\pi) \approx 0.9963$. Thus, the maximum phase after activation is significantly restricted. In contrast, the PhaseTanhshrink, peaks at $\pi - \tanh(\pi) \approx 2.1453$, allowing for a noticeably bigger phase. Still, both restrict the possible output space as the phase no longer spans the whole possible angle between the real part and imaginary vector of $[0, \pi]$.

\begin{figure*}[htb]
    \centering
    \begin{subfigure}[c]{0.49\textwidth}
        \centering
        \includegraphics{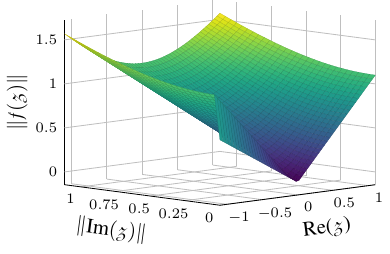}
        \caption{PhaseTanh Magnitude}
        \label{fig:PhaseTanh-mag}
    \end{subfigure}
    \hfill
    \begin{subfigure}[c]{0.49\textwidth}
        \centering
        \includegraphics{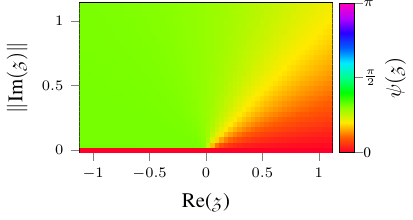}
        \caption{PhaseTanh Phase}
        \label{fig:PhaseTanh-phase}
    \end{subfigure}
    \begin{subfigure}[c]{0.49\textwidth}
        \centering
        \includegraphics{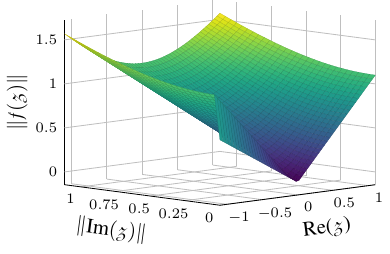}
        \caption{PhaseTanhshrink Magnitude}
        \label{fig:PhaseTanhshrink-mag}
    \end{subfigure}
    \hfill
    \begin{subfigure}[c]{0.49\textwidth}
        \centering
        \includegraphics{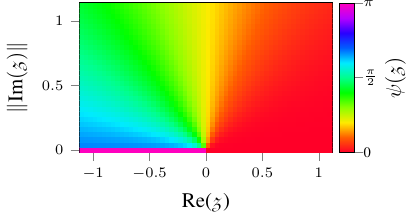}
        \caption{PhaseTanhshrink Phase}
        \label{fig:PhaseTanhshrink-phase}
    \end{subfigure}
    \caption{Visualization of the PhaseTanh and PhaseTanhshrink.}
    \label{fig:PhaseTanh-figures}
\end{figure*}

\begin{figure*}[htb]
    \begin{subfigure}[c]{0.49\textwidth}
        \centering
        \includegraphics{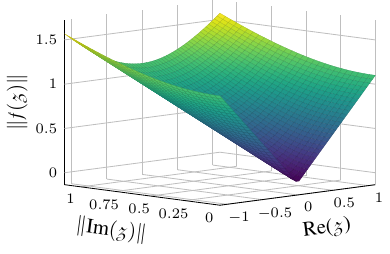}
        \caption{ScaledPhaseTanh Magnitude}
        \label{fig:NormalizedPhaseTanh-mag}
    \end{subfigure}
    \hfill
    \begin{subfigure}[c]{0.49\textwidth}
        \centering
        \includegraphics{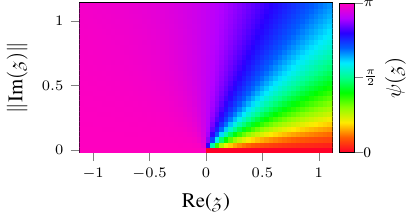}
        \caption{ScaledPhaseTanh Phase}
    \end{subfigure}
    \begin{subfigure}[c]{0.49\textwidth}
        \centering
        \includegraphics{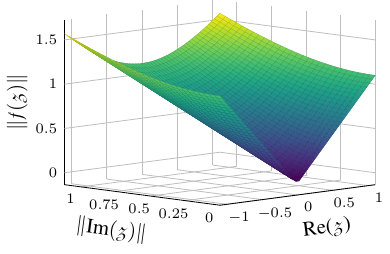}
        \caption{ScaledPhaseTanhshrink Magnitude}
        \label{fig:NormalizedPhaseTanhshrink-mag}
    \end{subfigure}
    \hfill
    \begin{subfigure}[c]{0.49\textwidth}
        \centering
        \includegraphics{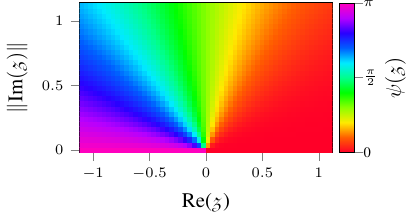}
        \caption{ScaledPhaseTanhshrink Phase}
    \end{subfigure}
    \caption{Visualization of the ScaledPhaseTanh and ScaledPhaseTanhshrink.}
    \label{fig:NormPhaseTanh-figures}
\end{figure*}

To address this issue, we further propose two adaptions: the ScaledPhaseTanh and the ScaledPhaseTanhshrink. The idea is to scale the angle after applying the non-linearity back to a range from $[0, \pi]$. Consequently, the phase is not restricted, but still non-linearly modified. The ScaledPhaseTanh is calculated by 
\begin{equation}
\begin{aligned}
    \phi_{\text{SPT}} (\quaternion{z}) 
    &= \lVert \quaternion{z} \rVert e^{\mathbf{n} \frac{\pi}{\tanh(\pi)} \tanh(\psi)} \\
    &= \lVert \quaternion{z} \rVert \left( 
        \cos\left(\frac{\pi \tanh(\psi)}{\tanh(\pi)}\right) + \frac{\mathbf{z}}{ \lVert \mathbf{z} \rVert} \sin\left(\frac{\pi \tanh(\psi)}{\tanh(\pi)}\right)
    \right)
\end{aligned}
\end{equation}
and the ScaledPhaseTanhshrink by 
\begin{equation}
\begin{aligned}
    &\phi_{\text{SPTs}} (\quaternion{z}) 
    = \lVert \quaternion{z} \rVert e^{\mathbf{n} \frac{\pi}{\tanhshrink(\pi)} \tanhshrink(\psi)} \\
    &= \lVert \quaternion{z} \rVert \left( 
        \cos\left(\frac{\pi \tanhshrink(\psi)}{\tanhshrink(\pi)}\right) + \frac{\mathbf{z}}{ \lVert \mathbf{z} \rVert} \sin\left(\frac{\pi \tanhshrink(\psi)}{\tanhshrink(\pi)}\right)
    \right) .
\end{aligned}
\end{equation}
The respective visualizations are shown in Figures \ref{fig:PhaseTanh-figures} and \ref{fig:NormPhaseTanh-figures}.

When observing Figures \ref{fig:PhaseTanh-mag} and \ref{fig:PhaseTanhshrink-mag} closely, we can see that for $\lVert \text{Im}(\quaternion{z}) \rVert = 0$ and $\text{Re}(\quaternion{z}) < 0$ there is a drop in the magnitude. This is caused by the fact that the imaginary vector vanishes and the magnitude is only determined by the real part, namely $\quatCompR{a} = \quatCompR{z}\cos(\tanh{(\pi})) \approx0.543\quatCompR{z}$ / $\quatCompR{a} = \quatCompR{z}\cos(\tanhshrink{(\pi})) \approx -0.543 \quatCompR{z}$. Thus, the real part is damped and the imaginary vector can't compensate in the limit $\lVert \text{Im}(\quaternion{z}) \rVert = 0$ as it does in other cases.
As indicated by Figures \ref{fig:NormalizedPhaseTanh-mag} and \ref{fig:NormalizedPhaseTanhshrink-mag}, the scaling prevents this behavior.

\paragraph{PhaseSin}
Finally, we propose the PhaseSin and ScaledPhaseSin.
The rational stems from the complexity of calculation for the above phase affecting activation functions. Hence, we aim to obtain an overall simpler Cartesian representation with fewer trigonometric functions that need to be evaluated.  We define the PhaseSin as
\begin{equation}
\begin{aligned}
    \phi_{\text{PS}} (\quaternion{z}) 
    &= \lVert \quaternion{z} \rVert e^{\mathbf{n} \sin(\psi)} \\
    &= \lVert \quaternion{z} \rVert \left( 
        \cos(\sin(\psi)) + \frac{\mathbf{z}}{ \lVert \mathbf{z} \rVert} \sin(\sin(\psi)) \right) . 
    %
    \label{equ:PhaseSin1}
\end{aligned}
\end{equation}

At first glance, it might seem unintuitive to calculate sin and cos of the outcome of a $\sin$ function itself. However, if we recall Equation \eqref{equ:angle_sin_representation} we observe that we can obtain the angle $\psi$ using the $\arcsin()$ function. As the $\sin$ as the chosen non-linearity is the inverse operation, we obtain our desired simplification. This is further particularly useful from a computational efficiency point of view as it lowers the amount of trigonometric functions required. Thus, 

\begin{equation}
\begin{aligned}
    \phi_{\text{PS}} (\quaternion{z}) 
    &\stackrel{\mathmakebox[\widthof{=}]{\eqref{equ:angle_sin_representation}}}{=}  \lVert \quaternion{z} \rVert \Bigg( 
        \cos\left(\sin\left(\arcsin\left(\frac{\lVert \mathbf{z} \rVert}{\lVert \quaternion{z} \rVert}\right)\right)\right) \\
        &\phantom{{}=}+ \frac{\mathbf{z}}{ \lVert \mathbf{z} \rVert} \sin\left(\sin\left(\arcsin\left(\frac{\lVert \mathbf{z} \rVert}{\lVert \quaternion{z} \rVert}\right)\right)\right) \Bigg) \\
    &=  \lVert \quaternion{z} \rVert \left( \cos\left( \frac{\lVert \mathbf{z} \rVert}{\lVert \quaternion{z} \rVert}\right) + \frac{\mathbf{z}}{\lVert \mathbf{z} \rVert} \sin\left( \frac{\lVert \mathbf{z} \rVert}{\lVert \quaternion{z} \rVert}\right)\right) .
    \label{equ:PhaseSin2}
\end{aligned}
\end{equation}

\begin{figure*}[htb]
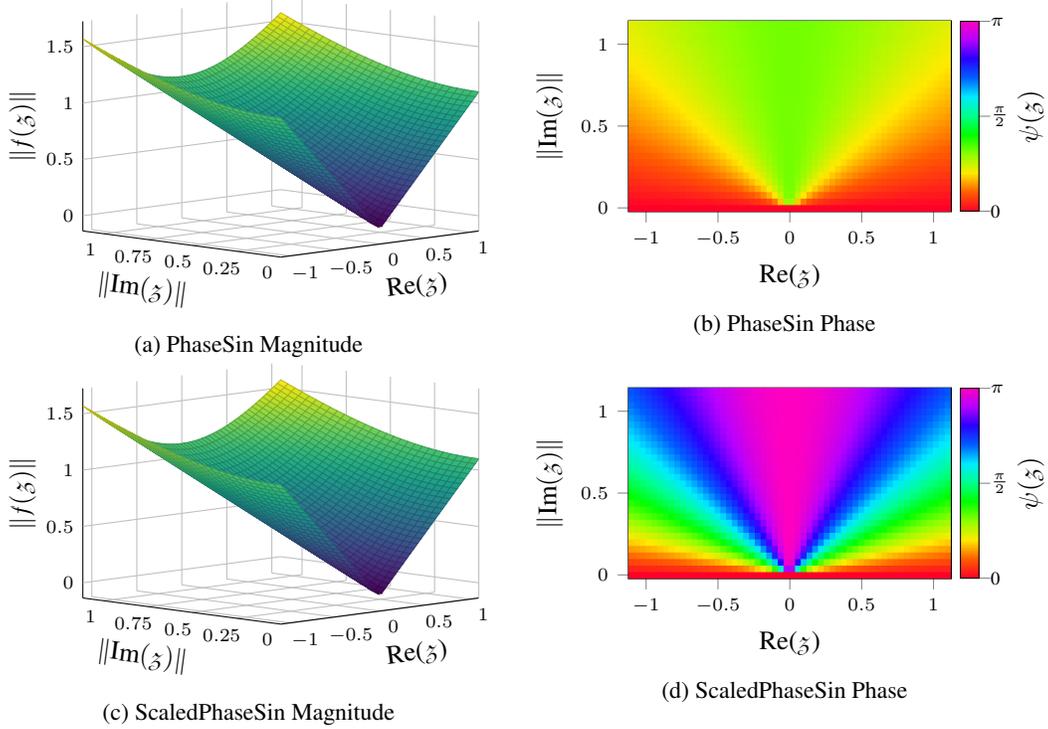

    \centering
    \begin{subfigure}[c]{0.49\textwidth}%
        \centering%
        \magPlot{act_visu_PhaseSin_psi_2.csv}
        \caption{PhaseSin Magnitude}%
    \end{subfigure}%
    \hfill%
    \begin{subfigure}[c]{0.49\textwidth}%
        \centering%
        \phasePlot{act_visu_PhaseSin_psi_2.csv}
        \caption{PhaseSin Phase}%
    \end{subfigure}
    \begin{subfigure}[c]{0.49\textwidth}%
        \centering%
        \magPlot{act_visu_NormPhaseSin_psi.csv}
        \caption{ScaledPhaseSin Magnitude}%
    \end{subfigure}%
    \hfill%
    \begin{subfigure}[c]{0.49\textwidth}%
        \centering%
        \phasePlot{act_visu_NormPhaseSin_psi.csv}
        \caption{ScaledPhaseSin Phase}%
    \end{subfigure}%
    \caption{Visualization of the PhaseSin and ScaledPhaseSin.}
    \label{fig:PhaseSin}
\end{figure*}

Consequently, the scaled version of the PhaseSin is calculated following
\begin{equation}
\begin{aligned}
    \phi_{\text{SPS}} (\quaternion{z}) 
    &= \lVert \quaternion{z} \rVert e^{\mathbf{n} \pi \sin(\psi)} \\
    &= \lVert \quaternion{z} \rVert \left( 
        \cos(\pi \sin(\psi)) + \frac{\mathbf{z}}{ \lVert \mathbf{z} \rVert} \sin( \pi \sin(\psi)) \right) \\
    &\stackrel{\mathmakebox[\widthof{=}]{\eqref{equ:angle_sin_representation}}}{=}  \lVert \quaternion{z} \rVert \Bigg( 
        \cos\left(\pi \sin\left(\arcsin\left(\frac{\lVert \mathbf{z} \rVert}{\lVert \quaternion{z} \rVert}\right)\right)\right) \\
        &\phantom{{}=}+ \frac{\mathbf{z}}{ \lVert \mathbf{z} \rVert} \sin\left(\pi \sin\left(\arcsin\left(\frac{\lVert \mathbf{z} \rVert}{\lVert \quaternion{z} \rVert}\right)\right)\right) \Bigg) \\
    &=  \lVert \quaternion{z} \rVert \left( \cos\left(\pi \frac{\lVert \mathbf{z} \rVert}{\lVert \quaternion{z} \rVert}\right) + \frac{\mathbf{z}}{\lVert \mathbf{z} \rVert} \sin\left(\pi \frac{\lVert \mathbf{z} \rVert}{\lVert \quaternion{z} \rVert}\right)\right) .
    \label{equ:ScaledPhaseSin}
\end{aligned}
\end{equation}
Again, the scaling ensures that the phase can span the whole range from $0 - \pi$ and is not capped at $\sin\left(\frac{\pi}{2}\right)=1$. The corresponding visualizations are shown in Figures \ref{fig:PhaseSin}.

\begin{remark}
    For the case where $\lVert \quatCompR{z} \rVert < 0$ we have to obtain the phase by calculating $\pi - \arcsin \left( \frac{\lVert \mathbf{z} \rVert}{\lVert \quaternion{z} \rVert} \right)$ since the maximum of the $\arcsin$ function is $\frac{\pi}{2}$. Hence, it does not cover the whole range $\psi(\quaternion{z}) = [0, \pi]$ the phase spans, just the area where the real part is non-negative. However, due to
    \begin{equation}
    \begin{aligned}
        &\sin \left( \pi - \arcsin \left( \frac{\lVert \mathbf{z} \rVert}{\lVert \quaternion{z} \rVert} \right) \right) \\
        &=
        \sin \left( \pi \right) \cos \left( \arcsin \left( \frac{\lVert \mathbf{z} \rVert}{\lVert \quaternion{z} \rVert} \right) \right) - 
        \sin \left( \arcsin \left( \frac{\lVert \mathbf{z} \rVert}{\lVert \quaternion{z} \rVert} \right) \right) \cos \left( \pi \right) \\
        &=
        0 * \cos \left( \arcsin \left( \frac{\lVert \mathbf{z} \rVert}{\lVert \quaternion{z} \rVert} \right) \right) - 
        \sin \left( \arcsin \left( \frac{\lVert \mathbf{z} \rVert}{\lVert \quaternion{z} \rVert} \right) \right) * (-1) \\
        &= \sin \left( \arcsin \left( \frac{\lVert \mathbf{z} \rVert}{\lVert \quaternion{z} \rVert} \right) \right) \\
        &= \frac{\lVert \mathbf{z} \rVert}{\lVert \quaternion{z} \rVert}
    \end{aligned}
    \end{equation}
    the simplification in Equations \eqref{equ:PhaseSin2} and \eqref{equ:ScaledPhaseSin} still holds.
\end{remark}

\section{Activation Derivatives}
\label{sec:derivatives}

\newcommand{\pd}[2]{\frac{\partial #1}{\partial #2}}

In this section, we calculate and visualize the derivatives for the proposed activation functions. This is of particular importance to understand their sensitivity areas as well as potential singularities or areas where they are not well-behaved. Furthermore, we use the derivatives and their visualization to evaluate criterion \ref{enum:gradient-flow} which is one key requirement for the desired activation function behavior.

We utilize the GHR calculus \cite{xu_enabling_2015} to calculate the derivatives following
\begin{equation}
    \pd{\quaternion{a}(\quaternion{z})}{\quaternion{z}} = 
    \frac{1}{4} \left( \pd{\quaternion{a}}{\quatCompR{z}} - \pd{\quaternion{a}}{\quatCompI{z}} \imagI - \pd{\quaternion{a}}{\quatCompJ{z}} \imagJ - \pd{\quaternion{a}}{\quatCompK{z}} \imagK \right) .
\label{equ:ghrDeriv}
\end{equation}

For more details on gradients and optimization in \ac{QNN} we refer to \cite{poppelbaum_time_2024}.

\begin{figure*}[htb]
    \centering
    \begin{subfigure}[c]{0.49\textwidth}%
        \centering%
		\includegraphics{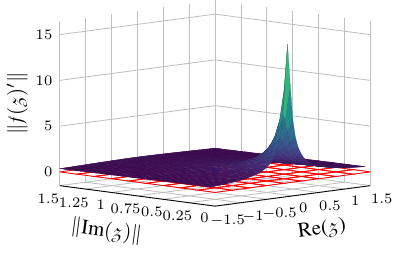}
        \caption{Norm}%
    \end{subfigure}%
    \hfill%
    \begin{subfigure}[c]{0.49\textwidth}%
        \centering%
		\includegraphics{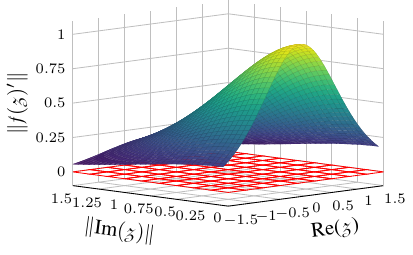}
        \caption{MagnitudeTanh}%
    \end{subfigure}
    \begin{subfigure}[c]{0.49\textwidth}%
        \centering%
		\includegraphics{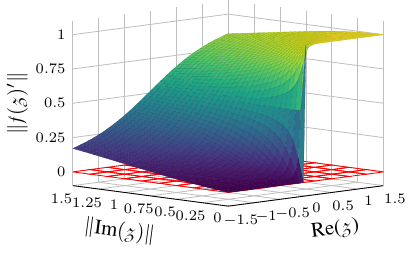}
        \caption{Cardioid}%
    \end{subfigure}%
    \caption{Visualization of the derivative of the activation functions affecting the magnitude. The red grid shows the plane where the z-value is zero.}
    \label{fig:act_derivatives_mag}
\end{figure*}

\begin{figure*}[htbp]
    \centering
    \begin{subfigure}[c]{0.49\textwidth}%
        \centering%
		\includegraphics{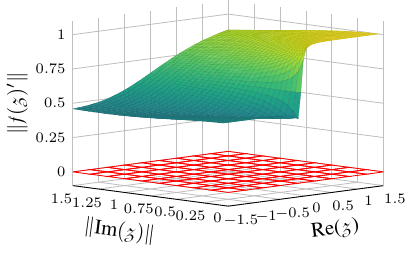}
        \caption{PhaseSin}%
    \end{subfigure}%
    \hfill%
    \begin{subfigure}[c]{0.49\textwidth}%
        \centering%
		\includegraphics{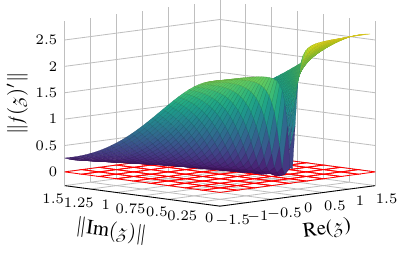}
        \caption{ScaledPhaseSin}%
    \end{subfigure}
    \begin{subfigure}[c]{0.49\textwidth}%
        \centering%
		\includegraphics{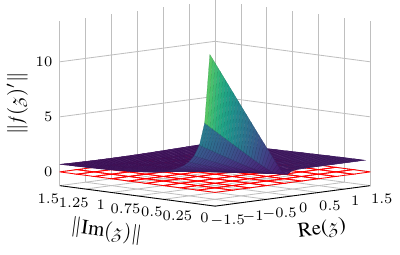}
        \caption{PhaseTanh}%
    \end{subfigure}%
    \hfill%
    \begin{subfigure}[c]{0.49\textwidth}%
        \centering%
		\includegraphics{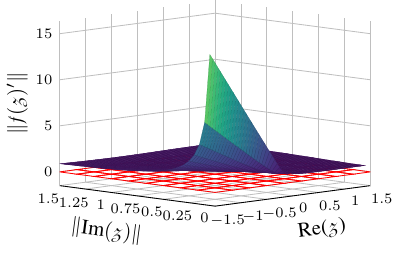}
        \caption{ScaledPhaseTanh}%
    \end{subfigure}
    \begin{subfigure}[c]{0.49\textwidth}%
        \centering%
		\includegraphics{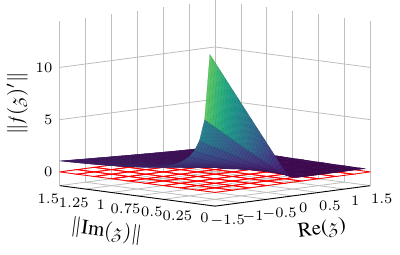}
        \caption{PhaseTanhshrink}%
    \end{subfigure}%
    \hfill%
    \begin{subfigure}[c]{0.49\textwidth}%
        \centering%
		\includegraphics{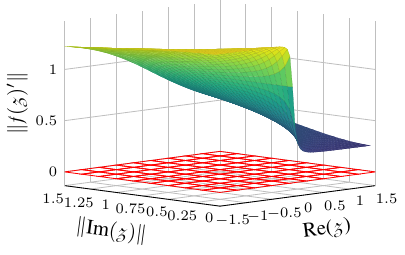}
        \caption{ScaledPhaseTanhshrink}%
    \end{subfigure}
    \caption{Visualization of the derivative of the activation functions affecting the phase. The red grid shows the plane where the z-value is zero.}
    \label{fig:act_derivatives_phase}
\end{figure*}

Using Equation \eqref{equ:ghrDeriv}, we obtained the derivatives listed in Table \ref{tab:derivatives}. We further visualized the respective derivative magnitudes in Figures \ref{fig:act_derivatives_mag} and \ref{fig:act_derivatives_phase}.

First of all, we can state that all activation functions are well suited to be used in \ac{QNN} due to their smooth gradient magnitudes without singularities and steadily high values. Thus, we don't expect any of the proposed functions to suffer from vanishing gradients caused by saturation effects.

\newcommand{\qV}[1]{\mathbf{#1}}
\newcommand{\norm}[1]{\lVert #1 \rVert}
\newcommand{\qnorm}[1]{\norm{\quaternion{#1}}}
\newcommand{\vnorm}[1]{\norm{\qV{#1}}}
\newcommand{\facTanhsrhink}{\frac{\pi}{\tanhshrink(\pi)}}

\begin{table*}[htb]
    \centering
    \caption{GHR derivatives for the proposed quaternion activation functions.}
    \label{tab:derivatives}
    \setlength{\tabcolsep}{3pt}
    \begin{tabular}{l l}
        \toprule
        Activation & GHR Derivative $\pd{\quaternion{a(\quaternion{z})}}{\quaternion{z}}$ \\
        \midrule
        Norm & 
        $\frac{3}{4 \lVert \quaternion{z} \rVert}$\\
        \addlinespace[2ex]MagnitudeTanh & 
        $
            \frac{3\tanh{\left( \lVert \quaternion z \rVert \right)}}{4\lVert \quaternion z \rVert }
            + \frac{1 - \tanh^2{\left( \lVert \quaternion z \rVert \right)}}{4}
        $
        \\
        \addlinespace[2ex]Cardioid &
        $
        \frac{1}{2} \left[ 1 + \frac{3\quatCompR{z}}{4\lVert \quaternion z \rVert} + \frac{\quaternion{z}}{4 \lVert \quaternion z \rVert} \right]$
        \\
        \addlinespace[2ex]PhaseSin &
        \makecell[l]{$
        \frac{1}{4}\Bigl[ \cos\left( \frac{\lVert \mathbf{z} \rVert}{\lVert \quaternion{z} \rVert} \right) \!
        \left(
        1 - 
        \frac{\mathbf{z}\quatConj{z}}{\lVert \quaternion{z} \rVert^2}
        + \frac{\quatConj{z}}{\lVert \quaternion{z} \rVert}
        \right) + 
        \sin\left( \frac{\lVert \mathbf{z} \rVert}{\lVert \quaternion{z} \rVert} \right) \!
        \left( 
        \frac{\quatConj{z}\lVert \mathbf{z} \rVert}{\lVert \quaternion{z} \rVert^2}
        + \frac{\mathbf{z}}{\lVert \mathbf{z} \rVert}
        + \frac{2\lVert \quaternion{z} \rVert}{\lVert \mathbf{z} \rVert}
        + \frac{\mathbf{z}\quatConj{z}}{\lVert \mathbf{z} \rVert\lVert \quaternion{z} \rVert}
        \right) \! \Bigr]
        $}\\
        \addlinespace[2ex]\makecell[l]{Scaled\\PhaseSin} &
        \makecell[l]{$
        \frac{1}{4}\Bigl[\cos\left( \frac{\pi \lVert \mathbf{z} \rVert}{\lVert \quaternion{z} \rVert} \right)\!
        \left(
        \pi \!-\! 
        \frac{ \pi \mathbf{z}\quatConj{z}}{\lVert \quaternion{z} \rVert^2}
        \!+\! \frac{\quatConj{z}}{\lVert \quaternion{z} \rVert}
        \right) \!+ 
        \sin\left( \frac{\pi \lVert \mathbf{z} \rVert}{\lVert \quaternion{z} \rVert} \right)\!
        \left( 
        \frac{\pi \quatConj{z}\lVert \mathbf{z} \rVert}{\lVert \quaternion{z} \rVert^2}
        \!+\! \frac{\pi \mathbf{z}}{\lVert \mathbf{z} \rVert}
        \!+\! \frac{2\lVert \quaternion{z} \rVert}{\lVert \mathbf{z} \rVert}
        \!+\! \frac{\mathbf{z}\quatConj{z}}{\lVert \mathbf{z} \rVert\lVert \quaternion{z} \rVert}
        \right)\!\Bigr]
        $}\\
        \addlinespace[2ex]PhaseTanh &
        \makecell[l]{$
        \frac{1}{4} \Bigl[ \cos{\left( \tanh(\psi) \right)} \left( 
            \frac{ -\quatCompR{z} (\tanh^2(\psi) - 1)}{\qnorm{z}}
            + \frac{\qV{z} (\tanh^2(\psi) - 1)}{\qnorm{z}}
            + \frac{\quatConj{z}}{\qnorm{z}}
        \right) + $\\$
        \sin{\left( \tanh(\psi) \right)} \left( 
            \frac{\quatCompR{z} \qV{z} (1 - \tanh^2(\psi))}{\vnorm{z} \qnorm{z}}
            + \frac{\qV{z} \quatConj{z}}{\vnorm{z}\qnorm{z}}
            + \frac{\vnorm{z} (1 - \tanh^2(\psi))]}{\qnorm{z}}
            + \frac{2\qnorm{z}}{\vnorm{z}}
        \right)\Bigr]
        $}\\
        \addlinespace[2ex]\makecell[l]{Scaled\\PhaseTanh} & 
        \makecell[l]{$
        \frac{1}{4} \Bigl[ \cos{\left( \frac{\pi \tanh(\psi)}{\tanh(\pi)} \right)} \left( 
            \frac{-\pi \quatCompR{z} (\tanh^2(\psi) - 1)}{\tanh(\pi) \qnorm{z}}
            + \frac{\pi \qV{z} (\tanh^2(\psi) - 1)}{\tanh(\pi) \qnorm{z}}
            + \frac{\quatConj{z}}{\qnorm{z}}
        \right) + $\\$
        \sin{\left( \frac{\pi \tanh(\psi)}{\tanh(\pi)} \right)} \Bigl( 
            \frac{\pi \quatCompR{z} \qV{z} (1 - \tanh^2(\psi))}{\tanh(\pi) \vnorm{z} \qnorm{z}}
            + \frac{\qV{z} \quatConj{z}}{\vnorm{z}\qnorm{z}}
            + \frac{\pi \vnorm{z} (1 - \tanh^2(\psi))]}{\tanh(\pi) \qnorm{z}}
            + \frac{2\qnorm{z}}{\vnorm{z}}
        \Bigr)\Bigr]
        $}\\
        \addlinespace[2ex]PhaseTanhshrink & 
        \makecell[l]{$
        \frac{1}{4} \Bigl[ \cos{\left( \tanhshrink(\psi) \right)} \left( 
            \frac{\quatCompR{z} \tanh^2(\psi)}{\qnorm{z}}
            - \frac{\qV{z} \tanh^2(\psi)}{\qnorm{z}}
            + \frac{\quatConj{z}}{\qnorm{z}} 
        \right) + $\\
        $\sin{\left( \tanhshrink(\psi) \right)} \left( 
            + \frac{\vnorm{z} \tanh^2(\psi)}{\qnorm{z}}
            + \frac{2\qnorm{z}}{\vnorm{z}}
            + \frac{\quatCompR{z} \qV{z} \tanh{(\psi)}^2}{\vnorm{z}\qnorm{z}} 
            + \frac{\qV{z}\quatConj{z}}{\vnorm{z}\qnorm{z}} 
        \right)\Bigr]
        $}\\
        \addlinespace[2ex]\makecell[l]{Scaled\\PhaseTanhshrink} & 
        \makecell[l]{$
        \frac{1}{4} \Bigl[ \cos{\left(\frac{\pi \tanhshrink(\psi)}{\tanhshrink(\pi)} \right)} \left( 
            \frac{\pi \quatCompR{z} \tanh^2(\psi)}{\tanhshrink(\pi) \qnorm{z}}
            - \frac{\pi \qV{z} \tanh^2(\psi)}{\tanhshrink(\pi)\qnorm{z}}
            + \frac{\quatConj{z}}{\qnorm{z}} 
        \right) + $\\$
        \sin{\left(\frac{\pi \tanhshrink(\psi)}{\tanhshrink(\pi)} \right)} \! \Bigl( 
            \frac{\pi \vnorm{z} \tanh^2(\psi)}{\tanhshrink(\pi) \qnorm{z}}
            + \frac{2\qnorm{z}}{\vnorm{z}}
            + \frac{\pi \quatCompR{z} \qV{z} \tanh{(\psi)}^2}{\tanhshrink(\pi)\vnorm{z}\qnorm{z}} 
            + \frac{\qV{z}\quatConj{z}}{\vnorm{z}\qnorm{z}} 
        \Bigr)\!\Bigr]
        $}\\
        \bottomrule
    \end{tabular}
\end{table*}

As expected, for the normalization we see the gradient rising significantly when $\text{Re}(\quaternion{z}) \rightarrow 0$ and $\lVert \text{Im}(\quaternion{z}) \rVert \rightarrow 0$, due to the norm of the input being in the denominator of the derivative. Thus, caution is required such that this does not impose issues into the training procedure. The MagnitudeTanh and Cardioid resemble the shape of the derivatives of their counterparts in $\mathbb{R}$ on the real axis. For the MagnitudeTanh we can observe that the sensitivity quickly drops when either $\text{Re}(\quaternion{z})$ or $\lVert \text{Im}(\quaternion{z}) \lVert$ increases. Similarly, the quaternion Cardioid has very low sensitivity for $\text{Re}(\quaternion{z}) < 0$ which only gradually increases with increasing $\lVert \text{Im}(\quaternion{z}) \rVert$ due to it's relationship with the ReLU activation.
Consequently, we don't see criterion \ref{enum:gradient-flow} as fulfilled for these activation functions. Nevertheless, the gradient magnitudes are still suitable for gradient based optimization.

For the activations affecting the phase, we can observe a very interesting property: They are sensitive basically everywhere and yield somewhat constant gradient magnitudes noticeably greater than zero as long as not $\lVert \text{Im}(\quaternion{z}) \rVert \rightarrow 0$.
Thus, they differ significantly from the previous discussed activation functions affecting the magnitude.
Especially the PhaseSin derivative stands out with the smoothest surface in the corresponding plot.
For the PhaseTanh, ScaledPhaseTanh and PhaseTanhshrink we can see a similar behavior to the norm, where there is a steep gradient increase when $\lVert \text{Im}(\quaternion{z}) \rVert \rightarrow 0$.
This is mainly caused by the term $\frac{2\qnorm{z}}{\vnorm{z}}$.
Thus, e.g. gradient clipping might be desired to include in the training routine such that this does not hurt the training process and stability.
Finally, if we compare the PhaseSin and ScaledPhaseSin derivatives with the PhaseTanh, ScaledPhaseTanh, PhaseTanshrink and ScaledPhaseTanhshrink derivatives is evident that the utilized simplifications also yields a way simpler derivative with less trigonometric functions to be evaluated, which is another argument in favor of these two proposed activation functions. In addition, they do not show any steep increase.

In Summary, especially due to the excellent sensitivity, we rate criterion  \ref{enum:gradient-flow} as fulfilled for all activation functions affecting the phase. Consequently, we expect improved gradient flow and low susceptibility to vanishing gradient problems, and thus improved performance of the \ac{NN} after training, which we further elaborate in the following section.

\section{Experimental Evaluation}
\label{sec:experiments}

In the following, we present the experimental evaluation of the proposed quaternion activation functions. Initially, we describe the experimental setup, followed by the results from our different conducted experiments.

\subsection{Experiment Setup}

For the experimental evaluation, we use the CIFAR-10 dataset \cite{Krizhevsky2009} 
as well as the SVHN dataset \cite{SVHN} since they are widely used in combination with \ac{QNN} \cite{Zhu_2018_ECCV, gai_reduced_2022, gaudet_deep_2018, grassucci_phnns_2022, vecchi_compressing_2020, ozcan_quaternion_2021}. 
Note that we do not aim for beating the latest benchmark results on these datasets, but instead aim for a comparison of the respective activation functions within a fixed training setup.
To form the quaternion valued inputs, the r-, g- and b-channels of the input image are stored in the $i$, $j$ and $k$ imaginary parts of the quaternion and the real part gets assigned all zeros. We train  with a batch-size of 256, an initial learning-rate of $1 \times 10^{-3}$ with an exponential learning rate decay to $1 \times 10^{-4}$ using the Adam optimizer \cite{kingma_adam} and the cross-entropy loss.
On CIFAR-10, we train for 100 epochs and on SVHN for 50 epochs.

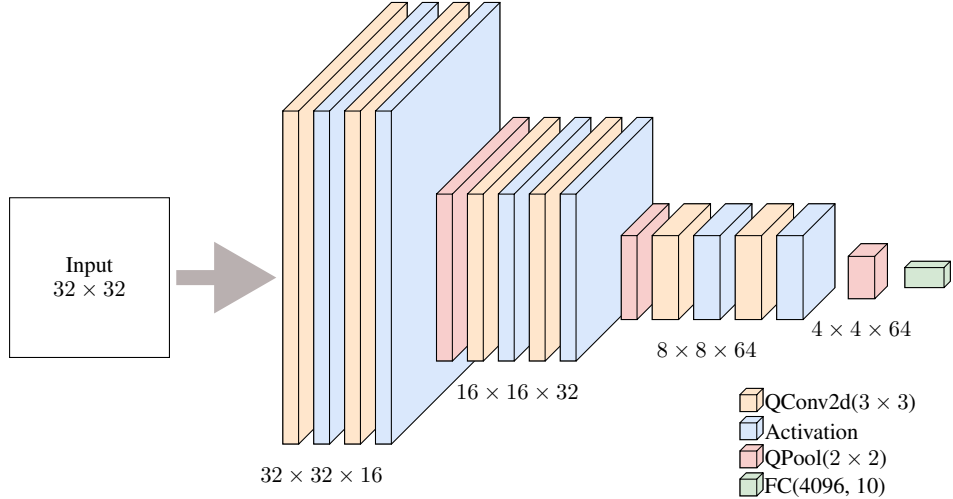
\begin{figure*}[htb]
    \centering%
%
%
%
%
%
\makeatletter
\pgfkeys{/pgf/.cd,
  parallelepiped offset x/.initial=2mm,
  parallelepiped offset y/.initial=2mm
}%
\pgfdeclareshape{parallelepiped}
{
  \inheritsavedanchors[from=rectangle] 
  \inheritanchorborder[from=rectangle]
  \inheritanchor[from=rectangle]{north}
  \inheritanchor[from=rectangle]{north west}
  \inheritanchor[from=rectangle]{north east}
  \inheritanchor[from=rectangle]{center}
  \inheritanchor[from=rectangle]{west}
  \inheritanchor[from=rectangle]{east}
  \inheritanchor[from=rectangle]{mid}
  \inheritanchor[from=rectangle]{mid west}
  \inheritanchor[from=rectangle]{mid east}
  \inheritanchor[from=rectangle]{base}
  \inheritanchor[from=rectangle]{base west}
  \inheritanchor[from=rectangle]{base east}
  \inheritanchor[from=rectangle]{south}
  \inheritanchor[from=rectangle]{south west}
  \inheritanchor[from=rectangle]{south east}
  \backgroundpath{
    \southwest \pgf@xa=\pgf@x \pgf@ya=\pgf@y
    \northeast \pgf@xb=\pgf@x \pgf@yb=\pgf@y
    \pgfmathsetlength\pgfutil@tempdima{\pgfkeysvalueof{/pgf/parallelepiped
      offset x}}
    \pgfmathsetlength\pgfutil@tempdimb{\pgfkeysvalueof{/pgf/parallelepiped
      offset y}}
    \def\ppd@offset{\pgfpoint{\pgfutil@tempdima}{\pgfutil@tempdimb}}
    \pgfpathmoveto{\pgfqpoint{\pgf@xa}{\pgf@ya}}
    \pgfpathlineto{\pgfqpoint{\pgf@xb}{\pgf@ya}}
    \pgfpathlineto{\pgfqpoint{\pgf@xb}{\pgf@yb}}
    \pgfpathlineto{\pgfqpoint{\pgf@xa}{\pgf@yb}}
    \pgfpathclose
    \pgfpathmoveto{\pgfqpoint{\pgf@xb}{\pgf@ya}}
    \pgfpathlineto{\pgfpointadd{\pgfpoint{\pgf@xb}{\pgf@ya}}{\ppd@offset}}
    \pgfpathlineto{\pgfpointadd{\pgfpoint{\pgf@xb}{\pgf@yb}}{\ppd@offset}}
    \pgfpathlineto{\pgfpointadd{\pgfpoint{\pgf@xa}{\pgf@yb}}{\ppd@offset}}
    \pgfpathlineto{\pgfqpoint{\pgf@xa}{\pgf@yb}}
    \pgfpathmoveto{\pgfqpoint{\pgf@xb}{\pgf@yb}}
    \pgfpathlineto{\pgfpointadd{\pgfpoint{\pgf@xb}{\pgf@yb}}{\ppd@offset}}
  }
}%
\makeatother
\tikzset{
  link/.style={
    draw,
    color=cust_grey,
    line width=2mm,
  },%
  conv/.style={
    parallelepiped,fill=white, draw,
    minimum width=0.1cm,
    minimum height=7.4cm,
    parallelepiped offset x=1.7cm,
    parallelepiped offset y=1.7cm,
    path picture={
      \draw[top color=drawio_orange_fill,bottom color=drawio_orange_fill]
        (path picture bounding box.south west) rectangle 
        (path picture bounding box.north east);
    },
    text=white,
  },%
  act/.style={
    parallelepiped,fill=white, draw,
    minimum width=0.1cm,
    minimum height=7.4cm,
    parallelepiped offset x=1.7cm,
    parallelepiped offset y=1.7cm,
    path picture={
      \draw[top color=drawio_blue_fill,bottom color=drawio_blue_fill]
        (path picture bounding box.south west) rectangle 
        (path picture bounding box.north east);
    },
    text=white,
  },%
  pool/.style={
    parallelepiped,fill=white, draw,
    minimum width=0.1cm,
    minimum height=7.4cm,
    parallelepiped offset x=1.7cm,
    parallelepiped offset y=1.7cm,
    path picture={
      \draw[top color=drawio_red_fill, bottom color=drawio_red_fill]
        (path picture bounding box.south west) rectangle 
        (path picture bounding box.north east);
    },
    text=white,
  },%
  fc/.style={
    parallelepiped,fill=white, draw,
    minimum width=0.6cm,
    minimum height=0.3cm,
    parallelepiped offset x=0.1cm,
    parallelepiped offset y=0.1cm,
    path picture={
      \draw[top color=drawio_green_fill, bottom color=drawio_green_fill]
        (path picture bounding box.south west) rectangle 
        (path picture bounding box.north east);
    },
    text=white,
  },%
}%
%
%
\begin{tikzpicture}[scale=0.85, every node/.style={scale=0.85}]%
    \node[conv, minimum height=5.2cm](conv1_1){};%
    \node[act, right=0.2cm of conv1_1, minimum height=5.2cm](act1_1){};%
    \node[conv, right=0.2cm of act1_1, minimum height=5.2cm](conv1_2){};%
    \node[act, right=0.2cm of conv1_2, minimum height=5.2cm](act1_2){};%
    \node[pool, right=0.6cm of act1_2, 
          minimum height=2.6cm, minimum width=0.1cm,
          parallelepiped offset x=1.2cm, parallelepiped offset y=1.2cm]
          (pool_1){};%
    \node[conv, right=0.2cm of pool_1, 
          minimum height=2.6cm, minimum width=0.2cm,
          parallelepiped offset x=1.2cm, parallelepiped offset y=1.2cm]
          (conv2_1){};%
    \node[act, right=0.2cm of conv2_1, 
          minimum height=2.6cm, minimum width=0.2cm,
          parallelepiped offset x=1.2cm, parallelepiped offset y=1.2cm]
          (act2_1){};%
    \node[conv, right=0.2cm of act2_1, 
          minimum height=2.6cm, minimum width=0.2cm,
          parallelepiped offset x=1.2cm, parallelepiped offset y=1.2cm]
          (conv2_2){};%
    \node[act, right=0.2cm of conv2_2, 
          minimum height=2.6cm, minimum width=0.2cm,
          parallelepiped offset x=1.2cm, parallelepiped offset y=1.2cm]
          (act2_2){};%
    \node[pool, right=0.6cm of act2_2, minimum height=1.3cm,
          minimum width=0.2cm,
          parallelepiped offset x=0.5cm, parallelepiped offset y=0.5cm]
          (pool_2){};%
    \node[conv, right=0.2cm of pool_2, minimum height=1.3cm,
          minimum width=0.4cm,
          parallelepiped offset x=0.5cm, parallelepiped offset y=0.5cm]
          (conv3_1){};%
    \node[act, right=0.2cm of conv3_1, minimum height=1.3cm,
          minimum width=0.4cm,
          parallelepiped offset x=0.5cm, parallelepiped offset y=0.5cm]
          (act3_1){};%
    \node[conv, right=0.2cm of act3_1, minimum height=1.3cm,
          minimum width=0.4cm,
          parallelepiped offset x=0.5cm, parallelepiped offset y=0.5cm]
          (conv3_2){};%
    \node[act, right=0.2cm of conv3_2, minimum height=1.3cm,
          minimum width=0.4cm,
          parallelepiped offset x=0.5cm, parallelepiped offset y=0.5cm]
          (act3_2){};%
    \node[pool, right=0.6cm of act3_2, minimum height=0.65cm,
          minimum width=0.4cm,
          parallelepiped offset x=0.2cm, parallelepiped offset y=0.2cm]
          (pool_3){};%
    \node[fc, right=0.4cm of pool_3]{};%
    \node[below=0.2cm of act1_1]{$32 \times 32 \times 16$};
    \node[below=0.2cm of act2_1]{\quad$16 \times 16 \times 32$};
    \node[below=0.2cm of act3_1]{$8 \times 8 \times 64$};
    \node[below=0.2cm of pool_3]{$4 \times 4 \times 64$};
    \node[conv, below=1.0cm of conv3_2,
          minimum height=0.25cm, minimum width=0.25cm,
          parallelepiped offset x=0.1cm, parallelepiped offset y=0.1cm,
          label=right:QConv2d($3\times3$)]
          (conv_legend){};%
    \node[act, below=0.15cm of conv_legend,
          minimum height=0.25cm, minimum width=0.25cm,
          parallelepiped offset x=0.1cm, parallelepiped offset y=0.1cm,
          label=right:Activation]
          (act_legend){};%
    \node[pool, below=0.15cm of act_legend,
          minimum height=0.25cm, minimum width=0.25cm,
          parallelepiped offset x=0.1cm, parallelepiped offset y=0.1cm,
          label=right:QPool($2\times2$)]
          (pool_legend){};%
    \node[fc, below=0.15cm of pool_legend,
          minimum height=0.25cm, minimum width=0.25cm,
          parallelepiped offset x=0.1cm, parallelepiped offset y=0.1cm,
          label=right:{FC(4096, 10)}]
          (fc_legend){};%
	\node[draw, align=center, minimum height=2.5cm,	minimum width=2.5cm, left=1.5 of conv1_1](inp){Input\\$32\times32$};%
    \draw [-triangle 60,link, minimum size=0.3cm] ([xshift=0.1cm]inp.east) -- ([xshift=-0.1cm]conv1_1.west);%
\end{tikzpicture}%
%
%
    \caption{Used Quaternion VGG Small (QVGG-S) model architecture. The channels indicate quaternion channels here, hence the input image consists of one quaternion channel instead of the usual 3 RGB-channels. Note that in contrast to other common VGG visualizations, we explicitly show the activation function.}
    \label{fig:vgg_s}
\end{figure*}

As the first model, we use a smaller VGG version with quaternion valued layer which we call QVGG-S. 
The objective of this is to avoid overparameterized models which can make up for flaws in the respective activation function and mitigate the differences between them. 
The model has a relatively low parameter count with 0.33 million parameters. 
It consists of 16 layers in total from which 6 are convolution layers, 6 activations, 3 pooling layers and one final linear layer. Note that this linear layer is real valued, hence it further serves as a learnable mapping from $\mathbb{H} \rightarrow \mathbb{R}$. The architecture is visualized in Figure \ref{fig:vgg_s}.

To test deeper architectures with higher parameter counts we further use a QVGG11 architecture as well as a QVGG16, thus models widely used in the literature. Note that for the QVGG16 we do the same modifications to the final linear layer as in \cite{grassucci_phnns_2022} to allow for a fair comparison later on. 
For the QVGG11 and QVGG16, we also incorporate a gradient clipping by norm of 10 when training using the SVHN dataset as we found that to be beneficial for the training process. 
Contrary to the default VGG architectures \cite{Simonyan2014VeryDC}, we use quaternion average pooling as in Equation \ref{equ:quat_avg_pool} instead of max pooling.

We waive on using any form of regularization or image augmentation as we want to obtain insights about the effect of just using different activation functions.
All experiments are performed using PyTorch \cite{Paszke2019PyTorchAI}.

\subsection{Results}
\label{subsec:results}

In our experiments, we include the respective activation functions from Section \ref{sec:quaternion_activations} in our QVGG-S model and train them as described. Each training is performed 10 times to mitigate the effect of random parameter initialization. This yielded the results as reported in Table \ref{tab:accuracies} and visualized in Figure \ref{fig:accs_cifar10} and \ref{fig:accs_svhn}. We report the maximum achieved accuracy as well as the mean accuracy including the standard deviation to obtain insight about the susceptibility to bad parameter initialization.

\begin{table*}[htb]
    \centering
    \caption{Achieved Accuracies when comparing the different proposed activation functions on the QVGG-S model. Maximum accuracy is highlighted bold and second best is underlined.}
    \label{tab:accuracies}
    \begin{tabular}{l c c c c}
        \toprule
        & \multicolumn{2}{c}{CIFAR-10 Accuracy in \%} & \multicolumn{2}{c}{SVHN Accuracy in \%} \\
        \cmidrule(lr){2-3} \cmidrule(lr){4-5}
        Activation & Max  & Mean + Std & Max  & Mean + Std \\
        \midrule
        Norm	                        & 80.25	& 79.805 $\pm$ 0.335 & 93.266 & 93.048 $\pm$ 0.191 \\
        MagnitudeTanh	                & 77.39 & 76.374 $\pm$ 0.466 & 92.294 & 91.904 $\pm$ 0.247 \\
        Cardioid	                    & 79.39 & 78.682 $\pm$ 0.488 & 94.249 & 94.030 $\pm$ 0.116 \\ 
        $\text{PhaseTanh}_{\psi}$	    & \underline{82.03} & \underline{81.724 $\pm$ 0.388} & 94.084 & 93.636 $\pm$ 0.258\\
        $\text{PhaseTanhshrink}_{\psi}$	& 81.63 & 81.157 $\pm$ 0.323 & 94.211 & 93.973 $\pm$ 0.109\\
        ScaledPhaseTanh$_{\psi}$          & 81.70 & 80.990 $\pm$ 0.509 & \textbf{94.925}	& \underline{94.466 $\pm$ 0.304} \\
        ScaledPhaseTanhshrink$_{\psi}$    & 81.20 & 80.372 $\pm$ 0.497 & 93.673	& 93.423 $\pm$ 0.153 \\
        PhaseSin$_{\psi}$               & \textbf{83.05}	& \textbf{82.649 $\pm$ 0.248} & 94.138 & 93.727 $\pm$ 0.340 \\  
        ScaledPhaseSin$_{\psi}$       	& 81.12 & 80.262 $\pm$ 0.359 & \underline{94.876} & \textbf{94.545	$\pm$ 0.213} \\
        \midrule
        Split-ReLU                      & 77.66 & 76.097 $\pm$ 0.910 & 94.084 & 93.922 $\pm$ 0.135\\
        Split-Tanh                      & 71.59 & 70.725 $\pm$ 0.400 & 91.499 & 90.930 $\pm$ 0.281\\
        \bottomrule
    \end{tabular}
\end{table*}
%
%
\noindent
\begin{figure*}[htb]%
    \centering%
    \accPlot{plot_data_cifar10.csv}{50}{86}{-5}{105}%
    \caption{Accuracy trajectory on the test set. The thick lines indicate the mean between the 10 runs, the respective semi-transparent area indicates the range between minimum and maximum accuracy.}
    \label{fig:accs_cifar10}
\end{figure*}
\begin{figure*}[htb]
    \centering
    \accPlot{plot_data_svhn.csv}{84}{95}{-2.5}{52.5}%
    \caption{Accuracy trajectory on the test set. The thick lines indicate the mean between the 10 runs, the respective semi-transparent area indicates the range between minimum and maximum accuracy.}
    \label{fig:accs_svhn}
\end{figure*}

As we can see, all proposed activation functions exhibit good training behavior without any major outliers. For CIFAR-10, the quaternion activation functions utilizing the phase perform the best with all of them achieving over 80\% as the mean accuracy. The best performing one is the PhaseSin with 82.649\% mean accuracy, followed by the PhaseTanh with 81.724\%. The activations where the magnitude is modified perform slightly worse with the Norm achieving a mean accuracy of 79.805\%, the Cardioid achieving 78.682\% and 76.374\% for the MagnitudeTanh. The elementwise application of the ReLU and Tanh yielded the worst overall results with 76.097\% and 70.725\% respectively.

For SVHN, the best mean result is obtained by the ScaledPhaseSin with 94.545\%, followed by the ScaledPhaseTanh achieving 94.466\%, with the inverse order for the maximum obtained accuracy. Third best performance is obtained by the Cardioid and forth by the PhaseTanhshrink. The MagnitudeTanh, Norm, PhaseTanh, ScaledPhaseTanhsrink and PhaseSin place below the ReLU this time in terms of mean accuracy whereas split-Tanh again performs worst with 90.822\%.

For the QVGG11 and QVGG16 models, we kept the training setup the same, however we trained with just three different random initialization to match the setup of \cite{grassucci_phnns_2022} and obtained the results presented in Tables \ref{tab:accuraciesVgg11} and \ref{tab:accuraciesVgg16}.

\begin{table*}[htb]
    \centering
    \caption{Achieved Accuracies when comparing the different proposed activation functions on the QVGG11 model. Maximum accuracy is highlighted bold and second best is underlined.}
    \label{tab:accuraciesVgg11}
    \begin{tabular}{l c c c c}
        \toprule
        & \multicolumn{2}{c}{CIFAR-10 Accuracy in \%} & \multicolumn{2}{c}{SVHN Accuracy in \%} \\
        \cmidrule(lr){2-3} \cmidrule(lr){4-5}
        Activation & Max  & Mean + Std & Max  & Mean + Std \\
        \midrule
        Norm                            & 74.79 & 74.187 $\pm$ 0.520 & 92.598 & 92.284 $\pm$ 0.223 \\
        MagnitudeTanh	                & 66.45 & 63.417 $\pm$ 2.235 & 89.663 & 66.075 $\pm$ 32.846 \\
        Cardioid   	                    & 78.49 & 77.817 $\pm$ 0.551 & 92.705 & 43.960 $\pm$ 34.468 \\
        PhaseTanh$_{\psi}$	            & 81.68 & 81.497 $\pm$ 0.205 & 94.349 & 94.166 $\pm$ 0.202 \\
        PhaseTanhshrink$_{\psi}$	    & 82.70 & 81.717 $\pm$ 0.708 & 94.415 & 94.407 $\pm$ 0.006 \\
        ScaledPhaseTanh$_{\psi}$	    & \underline{83.63} & \underline{82.897 $\pm$ 0.577} & \textbf{95.290} & \textbf{94.892 $\pm$ 0.310} \\
        ScaledPhaseTanhshrink$_{\psi}$	& 79.57 & 78.897 $\pm$ 0.528 & 94.495 & 94.443 $\pm$ 0.050 \\
        PhaseSin$_{\psi}$	            & \textbf{84.07} & \textbf{83.363 $\pm$ 0.601} & \underline{94.876} & \underline{94.795 $\pm$ 0.111} \\
        ScaledPhaseSin$_{\psi}$	        & 82.37 & 79.810 $\pm$ 1.844 & 94.826 & 94.767 $\pm$ 0.045 \\
        \midrule
        Split-ReLU                      & 77.44	& 77.077 $\pm$ 0.375 & 93.423 & 44.199 $\pm$ 34.807 \\
        Split-Tanh                      & 56.65	& 55.573 $\pm$ 0.781 & 22.223 & 20.466 $\pm$ 1.242 \\
        \bottomrule
    \end{tabular}
\end{table*}

\begin{table*}[htb]
    \centering
    \caption{Achieved Accuracies when comparing the different proposed activation functions on the QVGG16 model. Maximum accuracy is highlighted bold and second best is underlined.}
    \label{tab:accuraciesVgg16}
    \begin{tabular}{l c c c c}
        \toprule
        & \multicolumn{2}{c}{CIFAR-10 Accuracy in \%} & \multicolumn{2}{c}{SVHN Accuracy in \%} \\
        \cmidrule(lr){2-3} \cmidrule(lr){4-5}
        Activation & Max  & Mean + Std & Max  & Mean + Std \\
        \midrule
        Norm                            & 75.49 & 75.390 $\pm$ 0.128 & 93.642 & 93.477 $\pm$ 0.122 \\
        MagnitudeTanh	                & 69.54 & 68.773 $\pm$ 0.609 & 91.172 & 88.464 $\pm$ 3.279 \\ 
        Cardioid  	                    & 82.77 & 81.943 $\pm$ 0.585 & 19.587 & 19.587 $\pm$ 0.000 \\ 
        PhaseTanh$_{\psi}$	            & 84.96 & 83.863 $\pm$ 0.789 & 95.010 &  94.968 $\pm$ 0.047 \\
        PhaseTanhshrink$_{\psi}$	    & \underline{85.33} & 83.560 $\pm$ 2.454 & 95.402 &  95.210 $\pm$ 0.173 \\
        ScaledPhaseTanh$_{\psi}$	    & 85.14 & \underline{84.680 $\pm$ 0.528} & 95.363 &  95.290 $\pm$ 0.052 \\
        ScaledPhaseTanhshrink$_{\psi}$	& 84.08 & 82.977 $\pm$ 1.168 & 95.425 & 95.255 $\pm$ 0.120 \\
        PhaseSin$_{\psi}$ 	            & \textbf{85.62} & \textbf{85.153}	$\pm$ 0.406	& \textbf{95.548} & \textbf{95.413 $\pm$ 0.100} \\ 
        ScaledPhaseSin$_{\psi}$	        & 78.99 & 78.230 $\pm$ 0.541 & \underline{95.471} & \underline{95.356 $\pm$ 0.082} \\
        \midrule
        Split-ReLU                      & 79.79	& 79.247 $\pm$ 0.523 & 19.587 & 19.587 $\pm$ 0.000 \\
        Split-Tanh                      & 63.42	& 62.920 $\pm$ 0.522 & 85.591 & 61.920 $\pm$ 30.003 \\
        \bottomrule
    \end{tabular}
\end{table*}

In general, we observe that the findings from the QVGG-S hold. The unbounded activations prove to yield better results, and more importantly, the proposed quaternion activation functions consistently outperform the split activations,  which we attribute to the fulfillment of criterion \ref{enum:quat_components}. Further, comparing within the quaternion activation functions, fulfilling criterion \ref{enum:preservation_magnitude} appears to be more beneficial for the model performance than criterion \ref{enum:preservation_ratios}.
This is supported by Criterion \ref{enum:gradient-flow} which also holds for the activation functions affecting the Phase.
Among these activation functions, the PhaseSin is most often successful, which we attribute to its incorporated simplifications, which are particularly evident for the backward phase.

On both datasets and for both models, the activation functions affecting the Phase with one exception all perform better than the ReLU elementwise, and that the MagnitudeTanh outperforms the split-Tanh, even though both end up at the bottom of the performance chart. Further, for CIFAR-10 the Cardioid outperformed the split-ReLU in both experiments. For the SVHN dataset, when using the QVGG11 the MagnitudeTanh, split-ReLU and split-Tanh appear to be susceptible to collapsing trainings, and similar results can be observed for the QVGG16 models, whereas the activation functions affecting the phase appear to be very robust against variations or collapsing trainings due to different random parameter initializations.

\subsection{Investigation of the two Angle definitions}

By the way quaternions are composed, this leaves room for two interpretations of the incorporated phase / angle. The first is the angle between the real part and the imaginary vector as in Equation \eqref{equ:quaternion_phase}, the second is the angle of the encoded rotation of a unit quaternion as in Equation \eqref{equ:quaternion_rotation_angle}. They differ by the factor of two, but in combination with the non-linearity this ends up in a potentially big difference. Note that in the previous Subsections, we used the angle $\psi$. In the following, we want to investigate the effect of using the second angle option $\theta$. For that, the definitions of the quaternion activation functions as in Section \ref{sec:quaternion_activations} can remain unchanged, just the angle $\psi$ is replaced by $\theta$, satisfying $\theta = 2 \psi$.
Specifically, we perform these experiments for the PhaseTanh, PhaseTanhshrink, ScaledPhaseTanh and ScaledPhaseTanhshrink with their alternative visualizations as shown in Figure \ref{fig:acts_other_angle} in the \ref{sec:appendix}. It might be tempting to also follow that strategy for the Cardioid, however then it loses the ability to resemble the real valued ReLU when $\lVert \text{Im}(\quaternion{z}) \rVert$ = 0, which is one of its key features.

By performing the same experiments using the QVGG-S with the same setup also with the angle $\theta$, we end up with the results as shown in Tables \ref{tab:angle_accuracies_cifar10} and \ref{tab:angle_accuracies_svhn}.

\begin{table*}[htb]
    \centering
    \caption{Achieved Accuracies from the QVGG-S when comparing the two angle possibilities on the CIFAR-10 dataset. Higher accuracy is highlighted bold.}
    \label{tab:angle_accuracies_cifar10}
    \begin{tabular}{l c c c c}
        \toprule
        & \multicolumn{2}{c}{Angle $\psi$} & \multicolumn{2}{c}{Angle $\theta$} \\
        \cmidrule(lr){2-3} \cmidrule(lr){4-5}
        Activation & Max  & Mean + Std & Max  & Mean + Std \\
        \midrule
        PhaseTanh	        & 82.03 & 81.724 $\pm$ 0.388 & \textbf{83.11} & \textbf{82.752 $\pm$ 0.231} \\
        PhaseTanhshrink	    & \textbf{81.63} & \textbf{81.157 $\pm$ 0.323} & 79.87 & 79.277 $\pm$ 0.437 \\
        ScaledPhaseTanh       & \textbf{81.70} & \textbf{80.990 $\pm$ 0.509} & 75.73 & 74.851 $\pm$ 0.890 \\
        ScaledPhaseTanhshrink & \textbf{81.20} & \textbf{80.372 $\pm$ 0.497} & 76.70 & 75.844 $\pm$ 0.335 \\
        \bottomrule
    \end{tabular}
\end{table*}

\begin{table*}[htb]
    \centering
    \caption{Achieved Accuracies from the QVGG-S in \% when comparing the two angle possibilities on the SVHN dataset. Higher accuracy is highlighted bold.}
    \label{tab:angle_accuracies_svhn}
    \begin{tabular}{l c c c c}
        \toprule
        & \multicolumn{2}{c}{Angle $\psi$} & \multicolumn{2}{c}{Angle $\theta$} \\
        \cmidrule(lr){2-3} \cmidrule(lr){4-5}
        Activation & Max  & Mean + Std & Max  & Mean + Std \\
        \midrule
        PhaseTanh	        & 94.084 & 93.636 $\pm$ 0.258 & \textbf{94.503} & \textbf{94.191 $\pm$ 0.250} \\
        PhaseTanhshrink	    & \textbf{94.211} & \textbf{93.973 $\pm$ 0.109} & 92.955 & 92.871 $\pm$ 0.065 \\
        ScaledPhaseTanh       & \textbf{94.925} & \textbf{94.466 $\pm$ 0.304} & 93.493 & 93.054 $\pm$ 0.397 \\
        ScaledPhaseTanhshrink & \textbf{93.673} & \textbf{93.423 $\pm$ 0.153} & 92.417 & 92.086 $\pm$ 0.196 \\
        \bottomrule
    \end{tabular}
\end{table*}

As we can see, for both datasets, the PhaseTanh achieves the better performance using the angle $\theta$. Contrary, the PhaseTanhshrink, ScaledPhaseTanh and ScaledPhaseTanhshrink benefit more from utilizing the angle $\psi$. Especially for the scaled activations on CIFAR-10, the difference is quite significant, and the disadvantages of using the angle $\theta$ are greater than the improvements to be gained when using $\psi$ for the PhaseTanh. 

To further increase understanding on this topic and obtain more thorough insights, we also performed these experiments with the QVGG16 model and hence a way deeper architecture, yielding the results in Tables \ref{tab:angle_accuracies_cifar10_vgg16} and \ref{tab:angle_accuracies_svhn_vgg16}.

\begin{table*}[htb]
    \centering
    \caption{Achieved Accuracies from the QVGG16 when comparing the two angle possibilities on the CIFAR-10 dataset. Higher accuracy is highlighted bold.}
    \label{tab:angle_accuracies_cifar10_vgg16}
    \begin{tabular}{l c c c c}
        \toprule
        & \multicolumn{2}{c}{Angle $\psi$} & \multicolumn{2}{c}{Angle $\theta$} \\
        \cmidrule(lr){2-3} \cmidrule(lr){4-5}
        Activation & Max  & Mean + Std & Max  & Mean + Std \\
        \midrule
        PhaseTanh	            & \textbf{84.96} & \textbf{83.863 $\pm$ 0.789} & 83.71 & 83.377 $\pm$ 0.328 \\
        PhaseTanhshrink	        & \textbf{85.33} & \textbf{83.560 $\pm$ 2.454} & 83.75 & 82.663 $\pm$ 1.014 \\
        ScaledPhaseTanh	        & \textbf{85.14} & \textbf{84.680 $\pm$ 0.528} & 75.38 & 74.507 $\pm$ 0.825 \\
        ScaledPhaseTanhshrink	& \textbf{84.08} & \textbf{82.977 $\pm$ 1.168} & 80.10 & 78.603 $\pm$ 1.315 \\
        \bottomrule
    \end{tabular}
\end{table*}

\begin{table*}[htb]
    \centering
    \caption{Achieved Accuracies from the QVGG16 in \% when comparing the two angle possibilities on the SVHN dataset. Higher accuracy is highlighted bold.}
    \label{tab:angle_accuracies_svhn_vgg16}
    \begin{tabular}{l c c c c}
        \toprule
        & \multicolumn{2}{c}{Angle $\psi$} & \multicolumn{2}{c}{Angle $\theta$} \\
        \cmidrule(lr){2-3} \cmidrule(lr){4-5}
        Activation & Max  & Mean + Std & Max  & Mean + Std \\
        \midrule
        PhaseTanh	        & 95.010 & \textbf{94.968 $\pm$ 0.047} & \textbf{95.137} & 94.945	$\pm$ 0.189 \\
        PhaseTanhshrink	    & \textbf{95.402} & \textbf{95.210 $\pm$ 0.173} & 94.841 & 94.690	$\pm$ 0.116 \\
        NormPhaseTanh	    & \textbf{95.363} & \textbf{95.290 $\pm$ 0.052} & 93.927 & 93.155	$\pm$ 0.708 \\
        NormPhaseTanhshrink	& \textbf{95.425} & \textbf{95.255 $\pm$ 0.120} & 94.637 & 94.301	$\pm$ 0.238 \\
        \bottomrule
    \end{tabular}
\end{table*}

Again we can observe a better performance for the activation functions utilizing the angle $\psi$ instead of $\theta$ . This time, also the PhaseTanh obtains the higher mean accuracy using $\phi$ for both datasets, just for SVHN the maximum accuracy is slightly higher for $\theta$.
Thus, we decide to use $\psi$ for the remainder of this paper.

\subsection{Comparison with the Literature}

To put our results into broader context, in the following we list other works utilizing \acp{QNN} and the same datasets. 
Table \ref{tab:compt_literature} provides an overview about them and their results. For our results, we only list the best ones and report either the mean accuracy + standard deviation or the maximum accuracy, depending on the work we compare with.
We emphasize that this work is mainly intended to show the differences between the different activation functions and not to beat the latest benchmarks. 
Specifically, the main information gain lies in comparing these activation functions in a setup where everything except for the activation, i.e. model, training setup etc., is exactly the same to provide a fair comparison, as in Subsection \ref{subsec:results}.

\begin{table*}[htbp]
    \centering
    \setlength\tabcolsep{4.0pt} 
    \caption{Overview of the achieved accuracies in \% from multiple works utilizing \ac{QNN}. For our results with QVGG16, we report the mean accuracy with standard deviation separated, for QVGG11 we report the maximum accuracy. Maximum is highlighted bold and second best is underlined.}
    \label{tab:compt_literature}
    \begin{tabular}{l l c c c}
        \toprule
        Architecture & Activation & Param. & CIFAR-10 & SVHN \\
        \midrule
        VGG16 \cite{grassucci_phnns_2022}  & ReLU & 15M  & 85.067 $\pm$ 0.765 & 94.364 $\pm$ 0.394 \\
        QVGG16 \cite{grassucci_phnns_2022} & ReLU & 3.8M & 83.997 $\pm$ 0.493 & 93.887 $\pm$ 0.292 \\
        \multirow{3}{*}{\makecell{QVGG16\\(ours)}}
        & ScaledPhaseTanh$_{\psi}$	        & 3.8M & \underline{84.680 $\pm$ 0.528} & 95.290 $\pm$ 0.052 \\
        & ScaledPhaseSin$_{\psi}$	        & 3.8M & 78.230 $\pm$ 0.541 & \underline{95.356 $\pm$ 0.082} \\
        & PhaseSin$_{\psi}$ 	            & 3.8M & \textbf{85.153 $\pm$ 0.406} & \textbf{95.413 $\pm$ 0.100} \\ 
        \midrule
        QVGG11 \cite{gai_reduced_2022}      & ReLU & - & 73.26 & 91.92 \\
        \makecell[l]{RQVGG11\!+\!BN \cite{gai_reduced_2022}} & ReLU & - & 85.04 & 95.59 \\
        \multirow{2}{*}{\makecell{QVGG11\\(ours)}}
        & ScaledPhaseTanh$_{\psi}$	        & 7.1M & \underline{83.63} & \textbf{95.290} \\
        & PhaseSin$_{\psi}$	                & 7.1M & \textbf{84.07} & \underline{94.876} \\
        \midrule
        \cite{Zhu_2018_ECCV}     & ReLU & -     & 77.78 & - \\
        \cite{vecchi_compressing_2020} & ReLU & 0.5M  & 73.43 & - \\
        \cite{gaudet_deep_2018}        & ReLU & 0.13M & 93.23 & - \\
        \cite{gaudet_deep_2018}        & ReLU & 0.93M & 94.56 & - \\
        QCapsule \cite{ozcan_quaternion_2021} & - & 0.189M & 86.08 & 95.37 \\
        \bottomrule
    \end{tabular}
\end{table*}

The main work which we want to use to compare our results to is \cite{grassucci_phnns_2022} as they also use quaternion valued VGG style models, providing the best comparability for us. It has to be noted, however, that they reduce the layer size in the final fully connected layers in comparison to the default VGG architectures.
To allow for a fair comparison parameterwise, we also build our QVGG16 architecture with the same reduced layer size in the linear layers, however we stick to using a quaternion average pooling as in the QVGG-S. 
For the CIFAR-10 dataset, the ScaledPhaseTanh and the PhaseSin were able to outperform the QVGG16 using the ReLU activation function from \cite{grassucci_phnns_2022} in terms of mean accuracy, the PhaseSin even outperformed the real valued VGG16 implementation from \cite{grassucci_phnns_2022} with $\sim4$ times more parameter. On the SVHN dataset, all proposed activation functions affecting the phase outperformed the VGG16 as well as the QVGG16.
Even for our QVGG-S with more than 10 times fewer parameters, some of the proposed activation functions were able to outperform the QVGG16 using a ReLU activation and in two cases even the real valued VGG16 with 15Mio parameters when using the SVHN dataset.

In \cite{gai_reduced_2022}, VGG11-style Reduced Biquaternion Convolutional Neural Networks (RQV-CNN) are proposed. However, they rely on batch normalization to achieve good results for a VGG11 style architecture.
They also provide a VGG11-QCNN implementation for comparison.
Note that the authors use a non-standard subsampling of the datasets and that we compare with their results for a $3 \times 3$ kernel, they also ran other kernel sizes.
We outperform all the achieved accuracies from their VGG11-QCNN implementation on both datasets with the majority of our proposed activation functions.
Further, for CIFAR-10 the PhaseSin and for SVHN the ScaledPhaseTanh is competitive with the RQCVV11 despite lacking the batch normalization layers.

Besides that, there are further works using different layers and/or model architectures.
In \cite{Zhu_2018_ECCV}, convolution layer based on pure quaternions and quaternion rotation are used.
The authors of \cite{vecchi_compressing_2020} investigate the effect of regularizing \ac{QNN} using a model with 5 convolution layers and additional dropout layers for CIFAR-10. 
Note that we report the best result for \cite{vecchi_compressing_2020} which puts our non-regularized model in a disadvantage.
Nevertheless, we are able to outperform both approaches, even with the QVGG-S architecture with just 0.33 million parameters. 
Furthermore, \cite{gaudet_deep_2018} uses ResNet style \acp{QNN} with two different depths and parameter counts, however these models incorporate additional batchnorm layer as well as residual connections such that this does not yield a fair comparison with our model architecture.
Finally, the authors of \cite{ozcan_quaternion_2021} propose a quaternion based version of the capsule networks and hence also a completely different model architecture. Nevertheless, for our QVGG16, the PhaseSin, ScaledPhaseSin,  PhaseTanhshrink and ScaledPhaseTanhshrink achieve a higher maximum accuracy on the SVHN dataset and the PhaseSin is competitive for CIFAR-10.

\subsection{Discussion}

We aim to rate the success of the respective activation functions by recalling the criteria from Subsection \ref{subsec:criterias} and checking whether they are fulfilled or not

First and foremost, all used activation functions introduce sufficient non-linearity in the \ac{QNN} models such that non-linear relationships are learned, which is evident from the results achieved. This is required for \ac{QNN} composed out of standard quaternion valued building blocks like fully connected or convolution layer, which can easily be checked by using the identity function as the activation function, contrary to e.g. \cite{qin_fast_2022-2} where custom quaternion valued architectures are non-linear by itself. Also, all used activation functions are suitable to be used for gradient based optimization during the training phase and can be incorporated in arbitrary \ac{QNN} architectures. Thus, all activation functions fulfil criteria \ref{enum:non-linearity}, \ref{enum:optim} and \ref{enum:arbitrary_architectures}.
They differ, however, for the remaining criteria, as displayed in Table \ref{tab:criterias}.

\begin{table*}[htb]
    \centering
    \setlength\tabcolsep{4.0pt} 
    \caption{Comparison of the activation functions based on the established criteria.}
    \label{tab:criterias}
    \begin{tabular}{lcccccccc}
    \toprule 
    &\makecell[l]{Proper\\gradient flow\\+ sensitivity}
    &\makecell[l]{Utilization of\\all quaternion\\components}
    &\makecell[l]{Preservation\\of ratios} 
    &\makecell[l]{Preservation\\of magnitude} 
    &\makecell[l]{Interpre-\\tability} \\
    \midrule
    Norm                    & \xmark & \cmark & \cmark & \xmark & \cmark \\
    MagnitudeTanh           & \xmark & \cmark & \cmark & \xmark & \cmark \\
    Cardioid                & \xmark & \cmark & \cmark & \xmark & \cmark \\
    PhaseTanh               & \cmark & \cmark & \xmark & \cmark & \cmark \\
    PhaseTanhshrink         & \cmark & \cmark & \xmark & \cmark & \cmark \\
    ScaledPhaseTanh         & \cmark & \cmark & \xmark & \cmark & \cmark \\
    ScaledPhaseTanhshrink   & \cmark & \cmark & \xmark & \cmark & \cmark \\
    PhaseSin                & \cmark & \cmark & \xmark & \cmark & \cmark \\
    ScaledPhaseSin          & \cmark & \cmark & \xmark & \cmark & \cmark \\
    \midrule
    Split-ReLU              & \xmark & \xmark & \xmark & \xmark & \xmark \\
    Split-Tanh              & \xmark & \xmark & \xmark & \xmark & \xmark \\
    \bottomrule
    \end{tabular}
\end{table*}

When observing the gradients and sensitivities as obtained in Section \ref{sec:derivatives}, we see clear advantages for the activation functions affecting the phase, yielding the fulfillment of criterion \ref{enum:gradient-flow}.
Furthermore, all quaternion activation functions utilize all quaternion components, contrary to the split approach, such that criterion \ref{enum:quat_components} is only fulfilled for them. Likewise, the split approach neither preserves the ratios between the quaternion components nor the quaternion magnitude. 
Consequently, criterion \ref{enum:preservation_ratios} is fulfilled by the Norm, the MagnitudeTanh and the Cardioid, and criterion \ref{enum:preservation_magnitude} by all quaternion activation functions affecting the phase.
Finally, for the split approach, it is hard to interpret how the non-linearity was introduced in the \ac{QNN} whereas both, modifying the magnitude or the phase is much more comprehensive. Note that we explicitly do not aim to interpret the pure calculation of the activation function but instead how the activation changes the respective quaternions as a whole. Thus, we see criterion \ref{enum:interpretability} only fulfilled for the proposed quaternion activation functions and not for the split approach.

In all the experiments conducted, the activation functions affecting the quaternion phase appeared to be the best performing ones. 
We mainly attribute that to Criterion \ref{enum:quat_components} and \ref{enum:gradient-flow}, and also Criterion \ref{enum:preservation_ratios} seems to be more beneficial than \ref{enum:preservation_magnitude}.
By construction, these activation functions don't saturate while simultaneously producing proper gradients over the whole input range which do not tend to zero as e.g. for the Tanh in the areas where it saturates. Thus, vanishing gradient problems are less likely to appear during training. Consequently, in addition to utilizing all quaternion components for the activation calculation which emphasizes the combined quaternion characteristics during the forward phase, we obtain a further benefit of the proposed activation functions during the backward phase.
However, these advantages come at the cost that during the backward phase a lot of trigonometric functions need to be evaluated, which is costly in comparison to e.g the simple max() operation of the ReLU. If that is too costly, the Cardioid becomes an interesting alternative with its way simpler implementation, thus better run time efficiency, and a performance almost on par in most of our experiments. 

Overall, there is no clear overall favorite that could be a default go to activation such that still testing is required, but that is not different from real valued models, and we strongly advocate for the usage of activation functions that affect the phase, based on our findings.

\section{Conclusion}
\label{sec:conclusion}

In this paper, we introduced novel quaterniary activation functions, intended to be used within arbitrary \ac{QNN} architectures. 
Specifically, we proposed the MagnitudeTanh, the quaternion Cardioid, the PhaseTanh, the PhaseTanhshrink, the ScaledPhaseTanh, the ScaledPhaseTanhshrink, the PhaseSin and finally the ScaledPhaseSin.
After visualizing them utilizing the real part and the norm of the imaginary parts we investigated the respective GHR derivatives. 
There, we could observe that the activation functions modifying the phase are sensitive over the whole input range, while simultaneously not being susceptible to saturation and potential vanishing gradients caused by this.

In our experiments, we compared our proposed activation functions with the elementwise applied Tanh and ReLU on an image classification task using the CIFAR-10 and SVHN dataset, and three different VGG models.
For both datasets, most of the proposed activation functions proved to be superior. In particular, modifying the phase of the respective input quaternions yielded the best performance. Also, we evaluated the effect of using two different angle definitions and concluded that using the angle between the real part and the imaginary vector provided better performance in the vast majority of cases, in comparison to the angle of rotation encoded in a unit quaternion. In conclusion, we see great potential for improving \ac{QNN} architectures by incorporating the proposed quaternion activation functions.

In future work, we plan on investigating the effect of using the quaternion activation functions on different tasks besides image classification, to obtain further insights about the behavior of our proposed quaternion activation functions. Also, we aim to investigate the impact of modifying the encoded axis $\mathbf{n}$ which is not touched yet. This imposes new challenges as the length of the axis needs to be preserved. Further, this causes an activation function where the real part is unchanged but just the three imaginary parts change, hence a different behavior as investigated in this paper.

\printbibliography

\appendix 
\section{Visualizations for activations using the angle $\theta$ instead of $\psi$}
\label{sec:appendix}
\begin{figure*}[htb]
	\centering%
	\begin{subfigure}[c]{0.49\textwidth}%
		\centering%
		\includegraphics{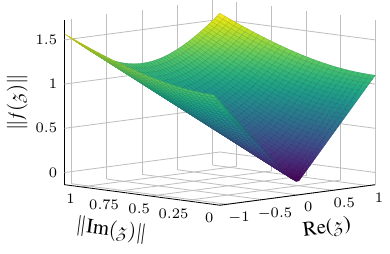}
		\caption{PhaseTanh$_{\theta}$ Magnitude}%
	\end{subfigure}%
	\hfill%
	\begin{subfigure}[c]{0.49\textwidth}%
		\centering%
		\includegraphics{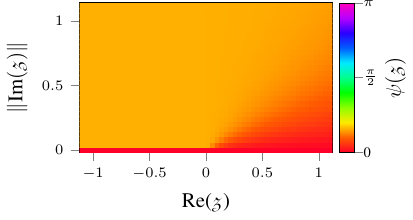}
		\caption{PhaseTanh$_{\theta}$ Phase}%
		\label{fig:acts_other_angle-PhaseTanh-phase}
	\end{subfigure}
	\begin{subfigure}[c]{0.49\textwidth}%
		\centering%
		\includegraphics{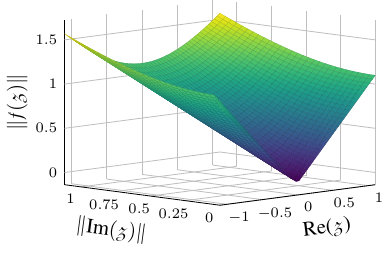}
		\caption{PhaseTanhshrink$_{\theta}$ Magnitude}%
	\end{subfigure}%
	\hfill%
	\begin{subfigure}[c]{0.49\textwidth}%
		\centering%
		\includegraphics{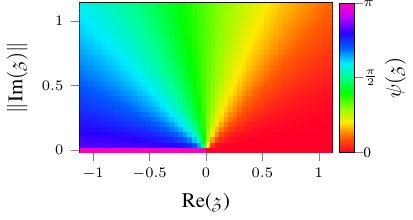}
		\caption{PhaseTanhshrink$_{\theta}$ Phase}%
		\label{fig:acts_other_angle-PhaseTanhshrink-phase}
	\end{subfigure}
	\begin{subfigure}[c]{0.49\textwidth}%
		\centering%
		\includegraphics{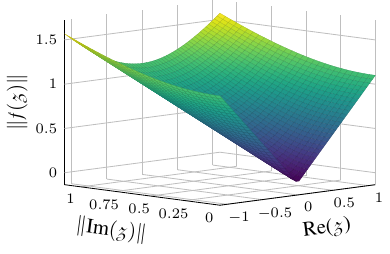}
		\caption{ScaledPhaseTanh$_{\theta}$ Magnitude}%
	\end{subfigure}%
	\hfill%
	\begin{subfigure}[c]{0.49\textwidth}%
		\centering%
		\includegraphics{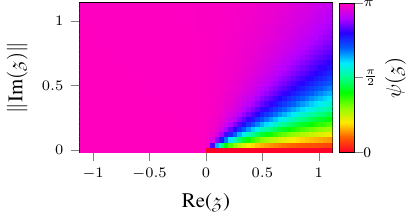}
		\caption{ScaledPhaseTanh$_{\theta}$ Phase}%
		\label{fig:acts_other_angle-NormPhaseTanh-phase}
	\end{subfigure}
\end{figure*}
\begin{figure*}[t!]
\ContinuedFloat
	\begin{subfigure}[c]{0.49\textwidth}%
		\centering%
		\includegraphics{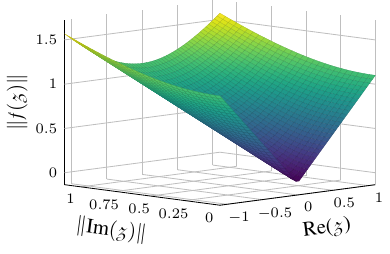}
		\caption{ScaledPhaseTanhshrink$_{\theta}$ Magnitude}%
	\end{subfigure}%
	\begin{subfigure}[c]{0.49\textwidth}%
		\centering%
		\includegraphics{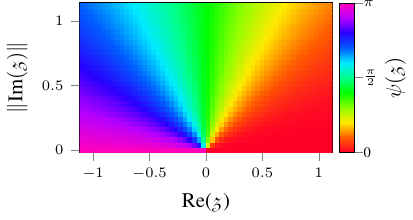}
		\caption{ScaledPhaseTanhshrink$_{\theta}$ Phase}%
		\label{fig:acts_other_angle-NormPhaseTanhshrink-phase}
	\end{subfigure}%
	\caption{Visualization of the PhaseTanh$_{\theta}$, PhaseTanhshrink$_{\theta}$, NormalizedPhaseTanh$_{\theta}$ and NormalizedPhaseTanhshrink$_{\theta}$.}%
	\label{fig:acts_other_angle}
\end{figure*}


\end{document}